\documentclass{article}

\usepackage{arxiv}

\usepackage[utf8]{inputenc} 
\usepackage[T1]{fontenc}    
\usepackage{hyperref}       
\usepackage{url}            
\usepackage{booktabs}       
\usepackage{amsfonts}       
\usepackage{nicefrac}       
\usepackage{microtype}      
\usepackage{lipsum}
\usepackage{graphicx}
\graphicspath{ {./images/} }
\usepackage{booktabs,
            makecell, 
            tabularx}
\usepackage{multirow}

\usepackage[english]{babel}
\usepackage{amsmath}
\usepackage{graphicx}
\usepackage{cite}
\usepackage{amsmath,amssymb,amsfonts}
\usepackage{algorithmic}
\usepackage{graphicx}
\usepackage[numbers]{natbib}
\usepackage{titlesec}
\usepackage{textcomp}
\usepackage{booktabs,
            makecell, 
            tabularx}

\usepackage[T1]{fontenc}    
\usepackage{url}            
\usepackage{booktabs}       
\usepackage{amsfonts}       
\usepackage{nicefrac}       
\usepackage{microtype}      
\usepackage{caption}
\usepackage{subcaption}
\usepackage{multirow}
\usepackage{epsfig}
\usepackage[colorinlistoftodos]{todonotes}

\title{Explainable AI for clinical risk prediction: a survey of concepts, methods, and modalities}

\author{
 Munib Mesinovic \\
  Department of Engineering Science\\
  University of Oxford\\
  Oxford, UK \\
  \texttt{munib.mesinovic@jesus.ox.ac.uk} \\
   \And
 Peter Watkinson \\
  Nuffield Department of Clinical Neurosciences\\
  University of Oxford\\
  Oxford, UK \\
  \And
 Tingting Zhu \\
  Department of Engineering Science\\
  University of Oxford\\
  Oxford, UK \\
}

\begin{document}
\maketitle
\begin{abstract}
Recent advancements in AI applications to healthcare have shown incredible promise in surpassing human performance in diagnosis and disease prognosis. With the increasing complexity of AI models, however, concerns regarding their opacity, potential biases, and the need for interpretability. To ensure trust and reliability in AI systems, especially in clinical risk prediction models, explainability becomes crucial. Explainability is usually referred to as an AI system's ability to provide a robust interpretation of its decision-making logic or the decisions themselves to human stakeholders. In clinical risk prediction, other aspects of explainability like fairness, bias, trust, and transparency also represent important concepts beyond just interpretability. In this review, we address the relationship between these concepts as they are often used together or interchangeably. This review also discusses recent progress in developing explainable models for clinical risk prediction, highlighting the importance of quantitative and clinical evaluation and validation across multiple common modalities in clinical practice. It emphasizes the need for external validation and the combination of diverse interpretability methods to enhance trust and fairness. Adopting rigorous testing, such as using synthetic datasets with known generative factors, can further improve the reliability of explainability methods. Open access and code-sharing resources are essential for transparency and reproducibility, enabling the growth and trustworthiness of explainable research. While challenges exist, an end-to-end approach to explainability in clinical risk prediction, incorporating stakeholders from clinicians to developers, is essential for success.
\end{abstract}


\section{Introduction}
\subsection{Motivation}
The advancement of artificial intelligence, and specifically machine learning and deep learning, in different areas of society, has been staggering. Applied AI can now beat humans in a string of complex logic-based games, can produce literary-level text, and enable a priori modelling of complicated protein structures that have long eluded biochemists \cite{wang2016does, schrittwieser2020mastering, dale2021gpt, jumper2021highly}. This progress has not eluded the field of healthcare where AI has seen progress at outperforming clinicians at breast cancer screenings, improved drug discovery, and predictions of several diseases, including, more recently, COVID-19 outcomes \cite{mckinney2020international, leibig2022combining, deng2022artificial, chassagnon2021ai, van2022critical}. With the rising capabilities of AI in patient care and prognosis come also rising risks of mismanagement, misclassification, misunderstanding (of the models themselves), and misuse. 

Greater prediction capabilities often come with greater complexities of models which makes them more opaque and unclear to both their developers as well as potential users. As far as clinical risk prediction models are concerned, decisions about patient treatment or diagnosis affect people’s lives profoundly, and AI systems are not without fault, sometimes leading to unreliable results or biased decision-making. For both clinicians and patients to understand and trust the decision-making process, especially when the impact of the decisions is significant, they need to be able to “evaluate and identify how [their personal] data is being used and whether the outcome is correct” \cite{branley2020user}. To do so, AI models are often checked for their explainability capacity which includes their interpretability and propensity to bias before being relied upon by clinicians. 

If potential harm is to be prevented, it is key to understand the inner workings of how these AI systems reach certain decisions. If prediction models are to be integrated into clinical practice, they inevitably need to do so based on high levels of reliability and soundness. Just one possible negative outcome of blindly implementing highly effective prediction models in the healthcare system is their amplification of existing biases and inequities \cite{dey2022human}. Clinicians can, often, identify noise or wrong features being picked up during learning by these models \cite{ghassemi2021false}. To fully unlock the impact of AI in healthcare, trust, interpretability, fairness, and bias all need to be addressed prior to deployment, and explainability plays a central role on this front.

Explainability, whose full definition will be addressed later, is an attribute of an (AI) automated decision system which describes the system's readiness to provide robust explanations of either its inner decision-making logic (inclusive) or the decisions themselves to human stakeholders \cite{arrieta2020explainable}. In other words, an explainable AI model can provide sufficient justification of its decisions, making it easier to identify potential flaws in learning and sources of bias. Explainability can also aid in promoting positive aspects of machine learning by providing new insights into predictive patterns relevant to disease prognosis and outcomes \cite{malhi2020explainable}. As highlighted in previous work, explainable AI applications can make more accurate predictions as well as offer increased transparency and fairness over their human counterparts in clinical applications \cite{goodman2017european}. These insights can help generalise the models to different patient populations, increasing robustness and decreasing chances of unequal treatment for protected groups, while also learning more about the model, the data, and the problem itself. 

Explainable AI models could, in addition, be a regulatory requirement. Existing regulatory frameworks like the General Data Protection Regulation (GDPR) might include a right to an explanation of automated decisions, depending on who one asks. Under Art. 15(1) the controller must provide “meaningful information about the logic involved” in AI systems, especially when explanations are needed to guarantee accuracy and to potentially challenge correctness \cite{voigt2017eu, selbst2018meaningful}. The interpretation of the word 'meaningful' leaves it to data protection authorities to decide on what information necessitates the enforcement of explainability. \cite{wachter2019right} are sceptical about legally enforced explainability since it seems that an explicit and legally sound right to an explanation was intentionally not included in the final version of the GDPR despite pressure from European lawmakers. Whatever the case may be, it is clear that future trajectories in AI applications to sensitive data areas like healthcare will lean towards robust explainability methods in clinical risk prediction. Finally, even when there is no legal obligation, it is important for clinicians to be able to both understand suggestions provided by clinical risk prediction models and be able to justify their own decision-making to colleagues and patients. 

This review hopes to present recent progress on multiple modality and methodology fronts in developing explainable models for clinical risk prediction. It also includes a summary assessment of these approaches with regard to quantitative and/or clinical evaluation and validation. 

\subsection{Contribution}

We aim to provide a comprehensive view of recent applied explainability work in clinical risk prediction which is inclusive of progress made across and within multiple modalities and which addresses topical concerns of reproducibility, external validation, and quantitative evaluation of explainability not attempted so far. Previous literature reviews, including \cite{loh2022application, nazar2021systematic, rasheed2022explainable, dey2022human, giuste2022explainable} have either been focussed on general classification and diagnosis using AI which is too broad of an approach, a specific modality, a specific disease, or more philosophical rather than implementation-based approaches to validation, fairness, trustworthiness etc. but never combining all of the above for a comprehensive commentary on the current trajectories of explainable AI in healthcare. A recently published review of explainable AI in \cite{dwivedi2023explainable} is relatively general and does not address the unique perspective and importance of explainable AI in domains like healthcare where they are of higher importance. Everything from stakeholders to taxonomy is seen through a new lense, and concepts like bias and trustworthiness must be addressed. While we do not aim for a social and/or critical analysis of concepts like trust, we do comment on the role these ideas play in achieving truly explainable AI for clinical risk prediction. We will reveal the lack of a common culture of clinically validating and quantitatively evaluating explainability methods in various modalities and applications for clinical risk prediction. We will also highlight patterns in the availability of reproducible code and results which undermines attempts at verifying research and increasing transparency of explainability work. In the end, we will highlight the need for a change in accepted research culture when developing and implementing existing or novel explainability frameworks that will answer a set of fundamental questions:

\begin{itemize}
    \item Does the proposed machine learning model for clinical risk prediction provide added clinical benefit whether in knowledge gain or practical importance?
    \item Can similar performance be achieved using a simpler or glass-box model instead of complex and costly architectures?
    \item Have the model prediction results been investigated by a clinician to some extent?
    \item Has the explainability method and its explanations for decisions been validated by a clinician or in a clinical setting?
    \item Has the explainability method been quantitatively evaluated in any manner, whether it be a metric or a comparative analysis?
    \item How will the explainable AI model benefit patients and how will that be clearly communicated?
\end{itemize}

\begin{figure}[h]
  \centering
  \includegraphics[width=0.8\linewidth]{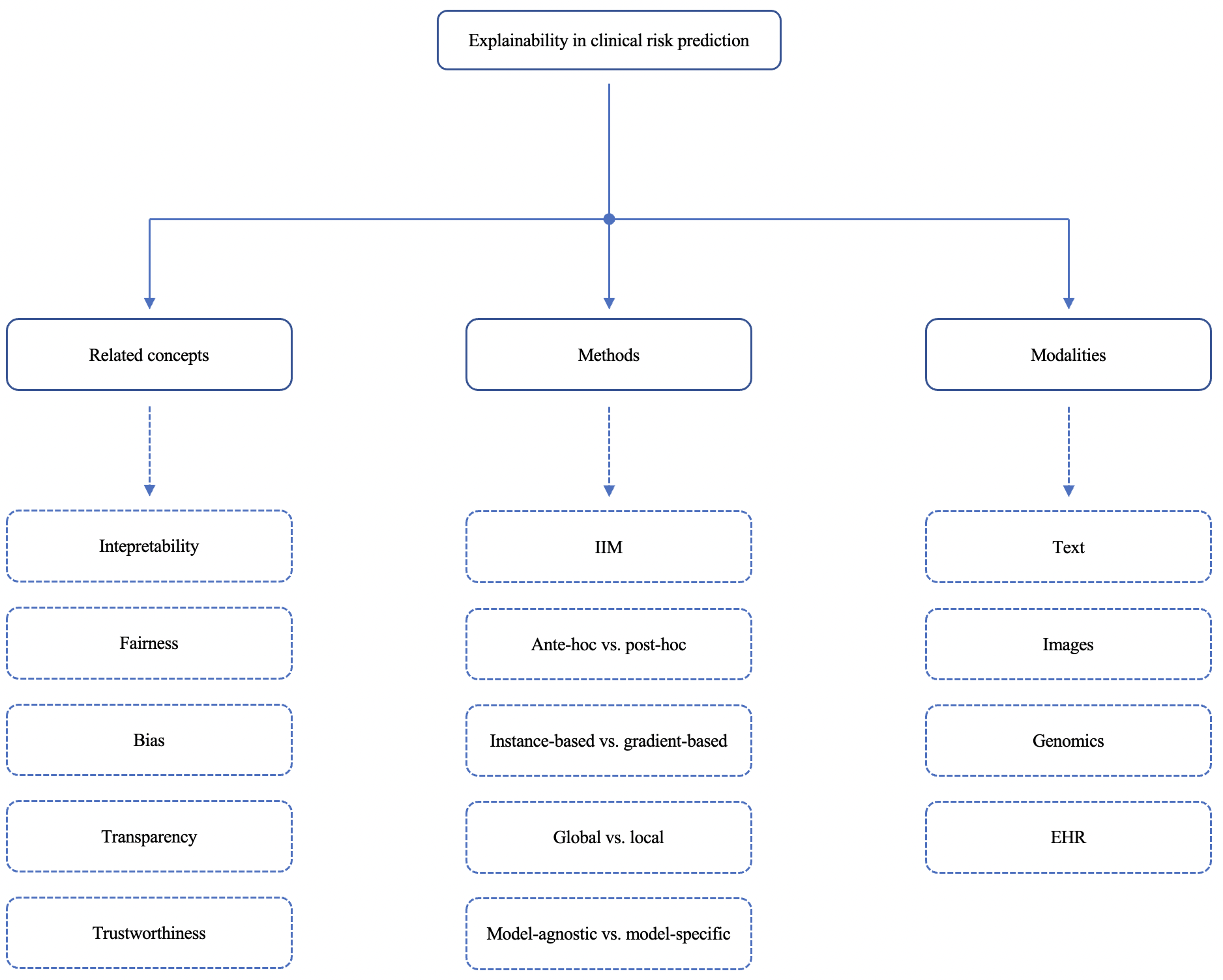}
  \caption{Framework for literature review structure showing the key terms, methods, and modalities investigated under XAI for clinical risk prediction.}
  \label{fig:Framework}
\end{figure}

The proposed framework for understanding the complex overlapping terms covered in this literature review can be seen in Figure \ref{fig:Framework} which encapsulates our reasoning of these topics in relation to explainability as a concept.

\section{Search Methodology}
\begin{figure}[h]
  \centering
  \includegraphics[width=0.8\linewidth]{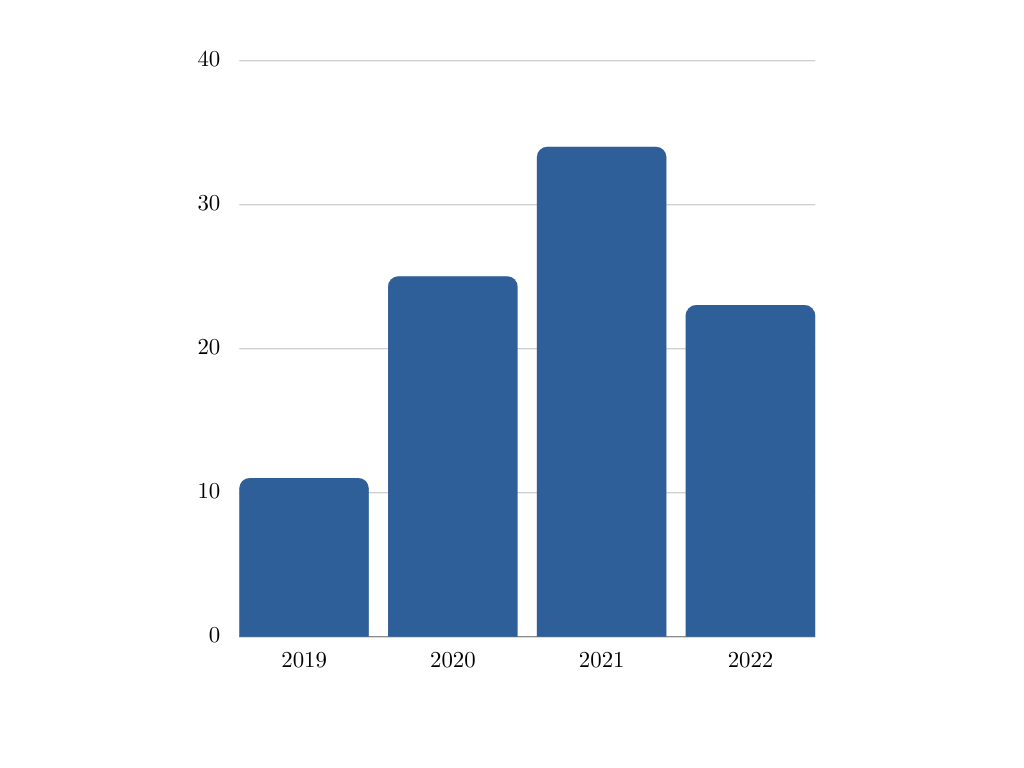}
  \caption{Frequency distribution of publications on XAI for clinical risk prediction since 2019 on PubMed database.}
  \label{fig:PubMed_citations}
\end{figure}

\begin{figure}[h]
  \centering
  \includegraphics[width=0.8\linewidth]{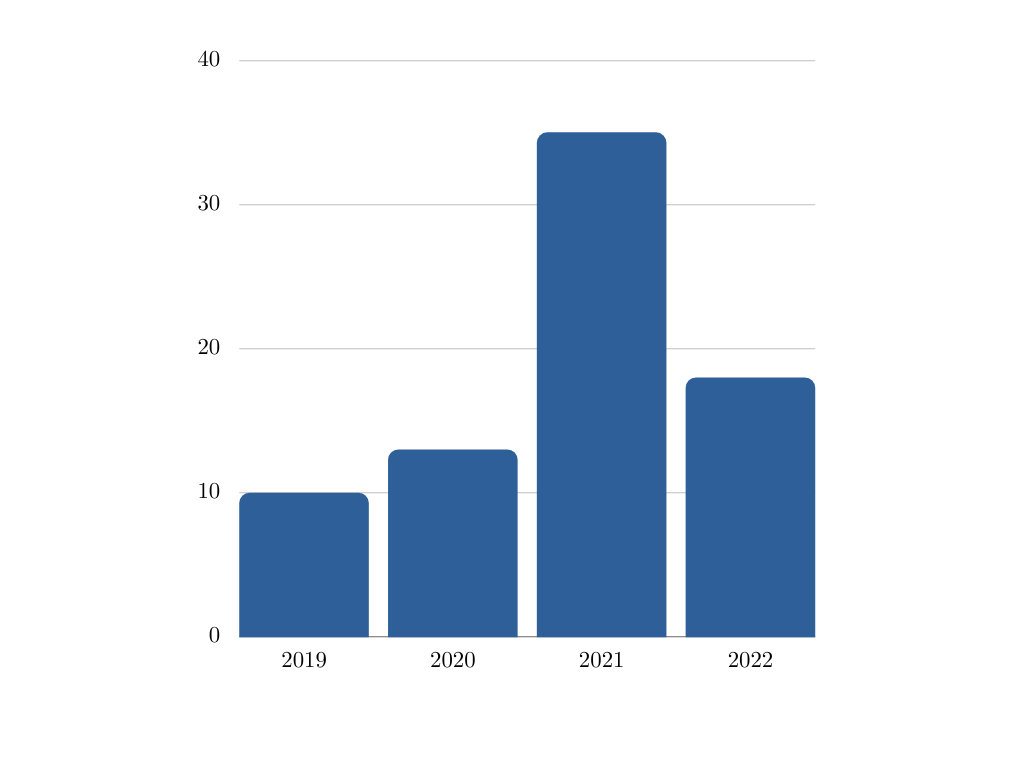}
  \caption{Frequency distribution of publications on XAI for clinical risk prediction since 2019 on IEEExplore database.}
  \label{fig:IEEExplore_citations}
\end{figure}

We used IEEExplore and PubMed to define the starting database search using the key terms identified in Table \ref{tab:key_terms_database}. For all of the databases, we considered only publications in the last 5 years which corresponds to the overwhelming majority of research on the topic due to its relatively recent development as can be seen in Figures \ref{fig:PubMed_citations} and \ref{fig:IEEExplore_citations}. Only papers published in English were reviewed. The final reference list was generated on the basis of originality and relevance to the scope of this Review by screening the titles and abstracts of the publications. For PubMed, we included publications classified as classical article, comparative study, evaluation study, mutlicenter study, observational study, technical report, and validation study. For IEEExplore, we considered only journal publications. There were 95 publications identified by PubMed and 76 by IEEExplore. Google Scholar was also perused for additional publications not included in the PubMed and IEEExplore search.

The publications were split into different groups depending on the data modalities concerned. A summary can be seen in Figure \ref{fig:Modalities}. One can see that text and EHR data occupy the largest portion of applications probably due to their relatively prevalent use in clinical risk prediction applications whereas imaging is more prevalent in diagnosis or classifications cases. The exclusion criteria included: 

\begin{figure}[h]
  \centering
  \includegraphics[width=0.8\linewidth]{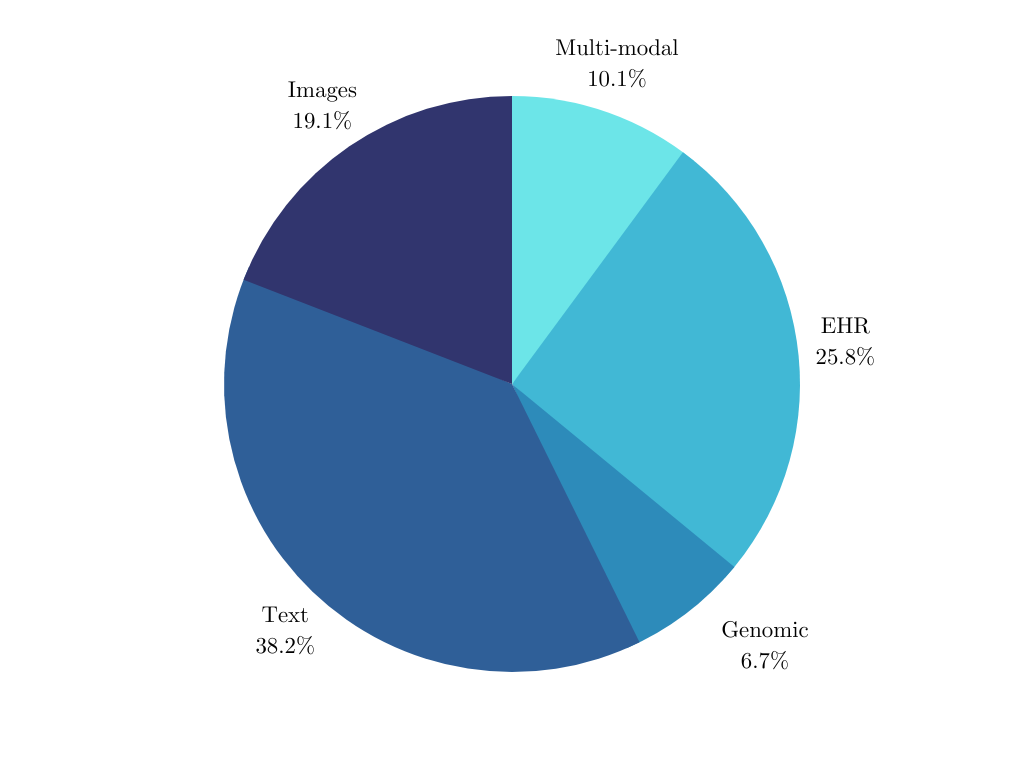}
  \caption{Pie chart of relative proportions of different modalities present in the explainability for clinical risk prediction applications literature.}
  \label{fig:Modalities}
\end{figure}

\begin{itemize}
    \item overlapping journal articles
    \item using the word interpretability in a clinical setting, for example when interpreting tumours from images without any relevance to machine learning interpretability per se
    \item works focused on classification and not clinical risk prediction
    \item journals not in the Q1 category of impact ranking
\end{itemize}

After applying the exclusion criteria we were left with a total of 89 publications to include in the review. \\

\begin{table}[h]
    \centering
    \caption{Search criteria for databases}
    \label{tab:key_terms_database}
    \setlength{\tabcolsep}{15pt}
    \begin{tabularx}{0.7\linewidth}{c
                                    c
                                    c
                                    c
                                 }
    
        \toprule
         "explainability" & "interpretability" & "XAI"  \\
          \midrule
          \textbf{AND} & & \\
          \midrule
          "machine learning" & "artificial intelligence" & "AI" \\
          \midrule
          \textbf{AND} & & \\
          \midrule
          "clinical"  & "medicine" & "healthcare" \\
          \midrule
          \textbf{AND} & & \\
          \midrule
          "prediction"  &  &  \\
            \bottomrule
    \end{tabularx}
\end{table}

\section{Background}
\subsection{Definitions and Evaluation Criteria}
Explainability in the context of AI refers to the ability to understand and not just interpret the decisions or predictions made by an AI system. A key terminological difference exists that must be made now between 'interpreting' and 'explaining' an AI model. To interpret a decision is to understand what it represents in the interpretable domain ie. in terms of the input features. Explaining a decision is not just functional but can also be deeper and more mechanistic in as far as it can concern the algorithmic underpinnings of the learning process of the model within \cite{montavon2018methods}. As such, explainability relates to the deeper inner working of the model while also concerning the broader social and human interactions of the model in ways that none of the individual concepts within do. As \cite{miller2019explanation} states, "explanations are social and involve conversations." From this philosophical perspective, interpretation is a subjective action, while explanation involves interactions \cite{zhong2021ai}. These interactions do not just include functional interpretations of results but the meaning of the results and how they impact humans (bias and fairness) as well as the comprehensibility of the model predictions to relevant stakeholders (transparency and trustworthiness). 

An important terminology note for the review is on the concept of clinical risk prediction versus diagnosis or classification. What we mean by AI for diagnosis or classification is in cases when the disease is usually already present and AI is used to help with faster, more accurate, or more insightful detection of the condition whereas risk prediction as we define it is more similar to prognostic clinical prediction, or predicting the risk of having the disease in the near- or long-term future. As such, prediction is then taken as a longitudinal instead of being a cross-sectional task and the patient population and associated data is more general with the first clinical starting point taken usually as inclusion criteria and the outcome occasionally being censored due to the existence of a prognostic time window and loss to follow-up \cite{van2021clinical}. This invites the use of survival analysis methods some of which are included in this review. Why that is important for explainability is that this allows for a greater focus on temporal trends and recurrent models for which some explainability methods have been developed but which under past systematic reviews have not received sufficient attention. For some modalities, however, like in the case of imaging, the boundary between clinical risk prediction and diagnosis is blurred especially since major explainability applications have been made on data rarely used for prognostic analysis. Inclusion of these research papers will be decided on a case-by-case basis even when explicit usage of clinical risk prediction is not specified so that a broader overview of explainability for medical imaging can be reliably ascertained. In other cases like for text data, clinical risk prediction is so rare that we decided to include applications to medical coding considering how sometimes medical coding applications can also be seen as examples of machine learning for phenotyping \cite{spasic2020clinical, shickel2017deep}. These cases will be kept to a minimum and they will not impact the overall analyses and conclusions of the literature review but rather are included for the sake of methodological completion of XAI applications.

In common practice across a myriad of machine learning research applications, the terms interpretability and explainability are often used interchangeably. And while interpretability is a key concept underpinning explainability, it is not fully equivalent to it, in fact, "there is no clear existing definition or evaluation criteria for interpretability" itself \cite{gilpin2018explaining}. More recent work has attempted to use metrics such as similarity, bias detection, execution time, and trust for measuring different interpretability methods' performance but the criteria are often subject to bias and might be measuring other aspects of the learning process not related to interpretability itself \cite{elshawi2021interpretability}. One way to attempt to define the concept is by describing its practical usefulness. For example, we can define interpretability as the degree to which humans can understand and comprehend the predictions made by machine learning models \cite{elshawi2021interpretability}. Here we go further and state that \textbf{interpretability of an AI system relates to how well the output predictions whether it be in clinical risk modelling or elsewhere can be interpreted by behaviour in the input features}. This can include concepts like feature importance and visualisations of learning \cite{gilpin2018explaining}. Following this line of thought, the "more interpretable a machine learning system is, the easier it is to identify cause-and-effect relationships within the system’s inputs and outputs" \cite{linardatos2020explainable}. An example could include a model learning to predict clinical risk of acute conditions like heart attack from time-series data and be able to highlight specific trends in time for measurement of heart rate important to making those predictions. Explainability, as we have mentioned earlier and will elucidate more ahead, is a wider concept which includes interpretability as it satisfies understanding internal learning logic and pattern recognition but is concerned also with the user experience of the system, data integration, problem motivation and definition, and bias and fairness.

The concept of bias has been increasingly mentioned in clinical risk prediction with AI influenced by similar challenges in other fields like law and insurance policy automation. Not all bias is necessarily problematic, as (machine) learning is often dependent on it, but certain patterns can be representative of underlying inequities and, in fact, propagate them further. \cite{cho2021rising} defines \textbf{bias as a kind of unfair systematic error that causes models to consistently wrongly predict for a certain subgroup of patients} and when those groups are vulnerable then bias can lead to unfair outcomes. Addressing bias is often done in a multi-faceted approach but it first requires detecting it. \cite{arrieta2020explainable} state that bias can be detected and corrected through explainability thereby implying that an explainable model would necessarily reduce bias or at the least expose it. Therefore, an explainable model whose every aspect, including data collection and integration, is understandable to human users such as clinicians, would be vulnerable to bias exposure and would be understandable to a level which would allow accounting for the bias in the system itself.

Fairness is related to bias which is often used interchangeably but which \cite{alelyani2021detection} note is more concerned with creating just usage of the systems rather than simply identifying bias in the data itself. \textbf{Bias is thus an aspect of the data or model learning whereas fairness or the lack of it comes as a result of machine decision-making under the influence of bias on vulnerable communities} \cite{mehrabi2021survey}. What this means in clinical risk modelling in practice is that both data need to be recognised for their potential biases as well as the feature processing and model inference stages, and the results need to be validated on various subpopulations either through counterfactual analysis by switching category groupings or stratified analysis etc. These interpretations and applications are up to clinicians and other users to make but they can only make them if the models are explainable enough to be able to understand the potential presence of unfairness.

Another example of fairness building up on bias rather than being equivalent terms are ensemble systems that check for bias as well as address its understanding, measurement, diagnosis and poential steps for mitigation \cite{ahn2019fairsight, bantilan2018themis}. Rather, it is not enough to state there is bias for a system to be evaluated for fairness, more steps need to be done to address it and communicate it. Several tools have been proposed to address these needs which consist of using several possible metrics for evaluating bias on the individual and global levels using distance, separation, and distortions in the feature and sample space. Feature correlations and perturbation analysis for sensitive clinical attributes can help identify measured biases. Mitigation can involve a range of steps including at the pre-, during, and post-processing stages. Machine learning models can be evaluated for fairness during learning with approaches like using counterfactual probabilities of two samples chosen from different groups being the same with respect to a given (non-sensitive) feature \cite{ahn2019fairsight, kusner2017counterfactual}. The end outcome of such tools are visualisations which reveal group skewing either through heatmaps or clusters and the impact what-if style prodding questions have on the machine learning model as part of the fairness framework. Of course, these methods rely on the assumption that the vulnerable or sensitive groups of samples and features are known a priori which is easier to specify in clinical scenarios due to extensive literature on bias and discrimination across clinical care. Another limitation is implied dichotomy in sensitive groups which might not always hold in cases of hierarchical systems of discrimination.

A system is considered transparent, as defined by \cite{das2020opportunities}, if it is "human-understandable." Transparency does not necessarily pertain to any internal methodology of the model, per se, but rather refers to the broader system encompassing the model, its design, motivation, development, distribution, and validation as well. Critical in the clinical realm, the end users are patients and/or clinicians whose trust must be depended on if these AI systems are to revolutionise healthcare practice \cite{preece2018stakeholders}. Transparency, as a requirement, thus, cannot just be achieved through the use of computational methods, rather, it relies on involving humans in the model creation pipeline end-to-end as they, and their experience, is what defined the system as transparent or not. Transparency, when satisfied, would allow patients, developers, and clinicians, as well as other stakeholders, to evaluate potential conflicts of interest and other negative consequences by opening up the system creation process outside of just developers' oversight \cite{chen2022explainable, char2020identifying, feudtner2018ethical}.

A common difference between transparency and interpretability, most of all, is that transparency pertains to parts outside the machine learning system itself. Interpretability concerns itself with interpreting the model prediction outputs as a function of the inputs. Transparency is focused on the user experience with the system as a whole \cite{arrieta2020explainable}. Per \cite{chen2022explainable}, a transparent AI system for healthcare needs to have its objective and clinical scenario elucidated and verified with clinicians and other stakeholders, including end users, before any system design has taken place. In other words, \textbf{transparency is characteristic of an AI system to have its motivation, design, and implementation elucidated to stakeholders whether it be through documentation or user-level testing or some other means}. Without explicitly stating how their published research has attempted to address the transparency requirements, researchers risk their proposals being inadequate for real-world implementations with clinical end users. 

Trust or trustworthiness is closely intertwined with the concept of transparency, as users tend to be more willing to rely on systems that have been developed end-to-end and applied in a transparent manner \cite{wolker2021algorithms}. Out of all of the concepts underpinning explainability, trustworthiness is the most difficult to separate, mostly because it is usually explained through a system already being fair, accountable, and transparent \cite{shin2020beyond}. Sometimes, trustworthiness is also defined as satisfying certain confidence levels in prediction results and the methods being interpretable \cite{ribeiro2016should}. For clinical risk prediction, a model must be sensitive, well-calibrated, and achieve good prediction results to achieve better trustworthiness, factors which are not accounted for in the other explainability sub-concepts. Therefore, to achieve integrated explainability, trustworthiness is the most linked concept with predictive performance, necessary for advancing change and implementing models in real-world clinical care \cite{lipton2018mythos}.

In the case of clinical risk prediction, here we propose the first-of-its-kind comprehensive evaluation review for peer-reviewed and published clinical risk prediction research which will comment on attempts or lack thereof of addressing the needs for explainability and what methods, validations, and open-access practices followed in achieving the same specifically focussed on clinical applications. To help aid the evaluation of this research we define some guiding questions in reviewing the papers:

\begin{itemize}
    \item Interpretability: Is there any attempt to comment on the most important features identified by the model during learning? Did the authors provide any analysis of the model learning so as to make the system more interpretable? What interpretability methods were applied and to what extent were they clinically validated?
    \item Fairness: Has bias been evaluated, diagnosed, visualised, communicated, or addressed?
    \item Bias: Did the authors comment on potential sources of bias for their system (including the data)? Are there any reflections on future risks of bias in the deployment or applications of the proposed model and its envisioned work scenarios?
    \item Transparency: Have the authors consulted clinicians or other experts as well as patients at any stage of the project process? Have they commented on their needs, motivations, and limitations as well as constraints regarding the model and its applications? Do the authors document and communicate these findings in their work?
    \item Trust: Have the model results been checked for random perturbations and noise (including the interpretability results)? What statistical or power tests were completed and do the authors indicate confidence intervals or standard errors in their results?
\end{itemize}

While each paper might not be evaluated for these questions, we will show how a simple check for clinical, statistical, and reproducibility tests show the current culture of explainable AI research for clinical risk prediction. In the review, different modalities for clinical usage will be considered, and each section will correspond to a separate modality under which a table will be included containing the summary information for each reference. Each row of the table will correspond to a specific reference and the underlying model, interpretability approach, dataset size, topic, and the existence of a clinical validation or open access check. quantitative evaluation will correspond either to simple statistical tests done on interpretability results and checking for noise as well as using several interpretability methods to confirm consistency of explanations. 

Clinical relevance is highly important in explainable AI for clinical risk prediction precisely because the ultimate goal of utilizing AI in healthcare is to improve patient outcomes and enhance clinical decision-making. Explainable AI models should not only be accurate and reliable but also provide insights and explanations that are meaningful and actionable for healthcare professionals. By emphasizing clinical relevance, explainable AI ensures that the generated explanations align with medical knowledge, guidelines, and the specific needs of healthcare practitioners. It enables clinicians to understand how the AI system arrives at its predictions or recommendations, thus building trust and confidence in the technology. This understanding allows clinicians to integrate AI outputs into their existing knowledge and expertise, resulting in more informed decision-making. 

Moreover, clinical relevance in explainable AI helps in the adoption and acceptance of AI systems within the healthcare domain. Healthcare professionals are more likely to trust and embrace AI technologies when the explanations provided by these systems are relevant and align with their clinical intuition. If the explanations are not clinically meaningful or comprehensible, clinicians may disregard or be skeptical of the AI system's recommendations, potentially leading to a lack of adoption and underutilization of valuable AI tools. Additionally, clinical relevance in explainable AI is crucial for patient-centered care. Patients and their families rely on healthcare professionals to make informed decisions about their health. When AI models provide explanations that are clinically relevant, it becomes easier for healthcare providers to communicate the reasoning behind their recommendations to patients. This transparency and shared decision-making process can enhance patient understanding, trust, and engagement in their own care. 

Explanations have no performance guarantees and, as such, the fidelity and value of their explanations is often insufficiently investigated, but in few cases of external human validation \cite{gilpin2018explaining}. Presenting a limitation in explainability work, inadequate external testing of explainability methods makes proposed explainable AI models for clinical risk prediction merely approximations of the model’s decision protocol and does not truly describe its underlying reasoning. Relying solely on using post-hoc explanations, as such, to assess the quality of model decisions could be just another source of error \cite{ghassemi2021false}. It is, thus, of high importance that while presenting an application or development of explainability methods to different modalities in clinical risk prediction, we also highlight their (non-)existent evaluations whether they be human in the form of clinical validation or algorithmic as the implementation of some quantitative metric. A relatively more common strategy that some could pursue is applying several different methods and comparing the explanations of each to guarantee either consistency or, under certain noise constraints, robustness. 

\subsection{Taxonomy of Methods}

\subsubsection{Inherently Interpretable Models (IIM)}
IIMs or, as they are often termed, glass-box or white-box models due to their relative decision-making transparency are among the simplest methods to achieve explainable prediction in clinical risk modelling. Their transparency is usually a consequence of the internal structure of the model itself being open to analysis and the relationship between input and output can be clearly represented either through equations or visualisations. Another key attribute for identifying IIMs is their lack of needing an external model or method to render the prediction model explainable \cite{rudin2019stop}. Entire groups of these methods have been proposed for various applications throughout time with some, as in the case of linear regression, preceding the development of machine learning itself.

Discretisation methods reduce a problem and its data into a set of discrete subsets thereby transforming the continuous solution space into one defined at discrete points. Decision trees and their ensembles resort to this strategy when learning by evaluating each feature value and selecting the cut-off to maximize the class separation as measured through entropy or Gini impurity (classification) or minimize the variance of the outcome (regression). Thus, each importance can be interpreted as share of the overall model importance \cite{molnar2020interpretable}. With its simple structure, it arguably provides more intuitive interpretability than sparse linear models whose outcomes are products of several layers of possibly non-linear functions in a multi-dimensional space. A limitation is that in cases of linear relationships between input and output, it approximates the behaviour through a series of discrete step functions which might not be of high fidelity.

Another example of discretisation methods are rule-based or expert-based systems, one of the earliest examples of domain adaptation of 'machine learning' to healthcare practice \cite{mccauley1992use}. These are methods that, similarly to decision trees, break down the outcome modelling into a series of discrete steps or rather decisions of an IF-THEN nature whose cut-off criteria are not determined by optimising a certain metric but from domain expertise or, in the case of Bayesian rule lists, from data mining the features for patterns \cite{letham2015interpretable}. As long as the decision rules are understandable, these models are radically interpretable, simple, and often times more compact than their decision tree cousins. Through domain expertise, they also apply a certain type of feature selection by \textit{a priori} defining what features are of importance from expert knowledge and only including those in the decision rule structure whereas linear models would assing weights to all features and learn a less sparse model. Similarly to decision trees, they are not adapted to linear modelling problems while, additionally, only being able to solve classification problems with discrete features. 

As compared to the previous methodologies, k-Nearest Neighbors provides an example- or instance-based interpretability whereas the earlier models account for global feature importance. Since it uses no modular learning of any global structure to make predictions there are also no global weights or decisions computer to interpret in the first place. The algorithm, in the case of classification, classifies the instance depending on the most common class in the neighbourhood of the data point in the feature space. Neighbourhood then depends on the number of neighbours and the distance metric one uses to identify the nearest k neighbours. To explain a specific prediction, one relies on the nearest neighbours used and their common feature patterns. Some work has used linear models to automate finding these patterns once clusters are formed \cite{zafar2021deterministic}. This implies, however, that in cases of higher dimensionality, seeing those patterns becomes intractable and, consequently, uninterpretable. 

Static or tabular data are not the only type of data encountered in clinical machine learning applications. Often, images or timeseries require a different \textit{modus operandi}, and deep learning, a machine learning paradigm relying on neural networks with a large number of layers (convolutional, recurrent, or otherwise), has been a great fit for these data formats and corresponding problems. Traditional machine learning models require feature engineering to create the input vectors from data, especially when they are images, whereas deep learning, through representation learning, allows the method to learn from raw data directly the optimal representation of internal features for learning \cite{lecun2015deep}. Causality and fuzzy logic can help make combined systems with deep learning \textit{inherently} interpretable. Traditional machine learning methods have seen their own implementations in the fuzzy logic realm but the focus here is on a more interesting approach of transforming inherently \textit{un}inerpretable deep learning otherwise by fuzzy rules. For decision tree, support vector machine, and nearest neighbour fuzzy implementations, consult the following literature: \cite{janikow1998fuzzy, marsala2009data, levashenko2007fuzzy, lin2002fuzzy, bian2020fuzzy}. The assumption in fuzzy logic is that instead of assuming input or output is deterministic, i.e. fixed value, one assigns uncertainty to both ends of the learning system. An example would be the decisions that a human makes while driving a car as provided by \cite{chimatapu2018explainable}. The decision-making of “if the distance to the car ahead is less than 2.5 m and the road is 10\% slippery then reduce car speed by 25\%” is approximated with numerical uncertainty reflecting the imprecise language of \textit{If} the distance to the car ahead is low and the road is slightly slippery \textit{Then} slow down. The numerical meanings of “low”, “close”, and “slow down” can vary and are, thus, fuzzy.

Breaking down the inference problem into fuzzy logic if-then steps aids interpretability and can be combined with high-performing opaque deep learning models through, for example, a fusion layer at the end of a neural network, one part of which is deep representational learning and the other having fuzzy rules \cite{deng2016hierarchical}. The weights in the fuzzy part of the networks are themselves fuzzy and can aid learning the joint probability distribution which, consequently, improves prediction performance. Another approach is to switch between neural and fuzzy layers or add the fuzzy logic system as a module with autoencoder structures through pre-trained layer replacement, often referred to as a deep type-2 fuzzy logic system (D2FLS) \cite{park2016intra, chimatapu2018interval}. Whatever the strategy, using fuzzy if-then rules holds great promise in overcoming the black-box nature of deep learning while keeping the superior predictive performance. There are limitations, however, as to how well suited these methods are for interpretability itself. Since they were built for uncertainty integration, they have not been yet optimised for explainability applications especially in higher dimensions \cite{chimatapu2020hybrid}. Future research can help address these concerns by implementing these methods in different case scenarios from static data processing to image classification and prediction while addressing the higher dimensionality through possible grouping of features.

The topic of using logic-based rules to interpret complex models without external add-ons necessarily also includes work on boolean rules generation with column generation methods (BRCGs). These methods attempt to reduce binary classification problems into a set of compact Boolean clauses similar to the ones seen earlier in rule-based models and fuzzy logic but with a set structure of more compact and easier to explain rules \cite{lakkaraju2016interpretable}. Each conjunction in these unordered rule sets is considered an individual rule and a positive prediction occurs when at least one of the rules is satisfied. These rules can be mined from the data (with decision trees for example) after which they can be selected through a myriad of methods including integer programming albeit which often comes with a limited search space \cite{dash2018boolean, wang2015learning}. Rule selection through column generation searches over an exponential number of all possible clauses optimised in a greedy manner with the combined reduced cost objective. This approach also allows a quantitative way of measuring intepretability by just looking at the size of the rules set itself. In cases of higher dimensionality, this approach becomes less tractable and is addressed by formulating an approximate Pricing Problem by randomly selecting a limited number of features and samples \cite{dash2018boolean}. The classification accuracy and computational costs for these binary classifiers are, however, a large limitation to their widespread acceptance as general IIM frameworks. Recent work has attempted to use optimised column generation to aid with the computational constraints and has included fairness into the objective evaluation through equality of opportunity and equalized odds constraints \cite{lawless2021fair}. 

\begin{figure}[h]
  \centering
  \includegraphics[width=0.8\linewidth]{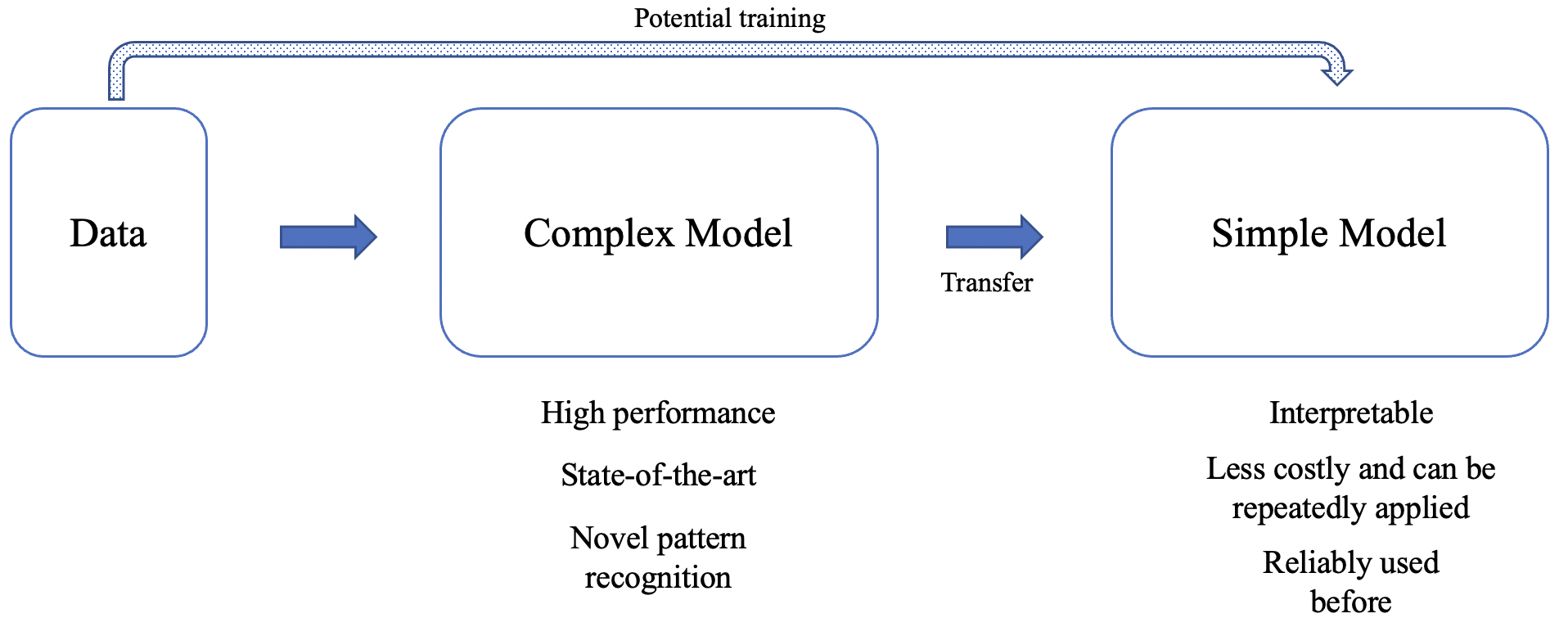}
  \caption{Using transfer learning to optimise the predictive performance of complex models including deep learning and transformers and combine them with simpler interpretable models that can be more reliably presented in clinical care.}
  \label{fig:Transfer}
\end{figure}

Besides relying on fusion layers and rule-based methods to turn opaque models inherently interpretable by some type of combination, another example that should be investigated more is incorporating transfer learning into explainability. The concept of transfer learning rests on pre-training or learning weights for one task and 'transferring' that knowledge and the learned weights to another problem, sometimes with a limited amount of re-learning and often to a different domain \cite{zhuang2020comprehensive}. By using transfer learning to preserve the high predictive performance of black-box models and transferring them to interpretable simpler models a more balanced trade-off between deep learning and transparency can be obtained. An abstract representation of this process can be seen in Figure \ref{fig:Transfer}. The limitation, however, is on how transferrable the features learned from the complex model are to the simpler model \cite{yosinski2014transferable}. Assuming sufficiently, through this approach, one can possibly simultaneously use a simpler and more reliable model that clinicians would be familiar with while preserving higher performance, apply a model trained on a large amount of data to a smaller dataset thereby avoiding overfitting, and reducing computational costs by simply 'cut-and-pasting' a pre-trained model to the problem at hand \cite{dey2022human, micaelli2019zero}. All of these can help achieve more explainable systems that are not just interpretable, but easier to check for fairness and bias, especially considering the smaller and more accessible datasets that can now be used for deployment. 

Methods like ProfWeight transfer the learning process from a pre-trained deep neural network with a high predictive capability to a simpler interpretable model or a very shallow network of low complexity and a priori low test accuracy. The method relies on flattened intermediate representations used to generate confidence scores of linear probes for weighting of samples during training of the simpler models. This approach has not yet been sufficiently evaluated in real-world applications such as clinical risk prediction but could hold great promise in achieving a stable balance between higher test accuracies and white-box models \cite{dhurandhar2018improving}. Possible extensions could include new and more challenging data domains and applications, experimenting with a combination of different models of varying complexity, and optimising for inference time and reducing costs since the initial weights still need to be computed from a complex deep learning model.

Inherently interpretable models also have their limitations when their explanations are often burdened by possible and unidentified confounders. Recent work on human agency with explainable AI has found that the added user reliance on using an IIM with less features and cleared decision-making is not significant. Additionally, researchers found that using an IIM limited users’ "abilities to detect when the model had made a sizable mistake, seemingly due to information overload caused by the amount of detail in front of them" \cite{poursabzi2021manipulating}. A key factor in usability then is the number of features included in the model itself. The tendency for outliers to break the explainability framework is to be addressed through collaborative explainability mechanisms between human users and automated systems by inviting users to give their own predictions before seeing the model. In this way, users might be more likely to notice any outlier values. Thus, while users might be more likely to rely on simpler and clearer models, the benefits to prediction accuracy are negligible, meaning that using IIMs with clearer features of a smaller number has first and foremost a psychological effect in building trust even if not a practical one. 

Recent advances, however, have been in the field of explainability for models and methods that do not fit the IIM criteria. Large, complex, and often opaque models that deal with learning patterns from text, images, genomics, and timeseries measurements present unique challenges when applied to clinical risk score prediction. Often times, separate explainability methodologies of similar complexity as the models they purport to explain have to be developed and tested. That will be the focus of the next section. 

\subsubsection{Ante-hoc vs. Post-hoc}

We have previously described several paradigms and methods underpinning inherently interpretable models in AI, including classic models in machine learning, such as decision trees and rule-based models. Often, because these models do not require any post-prediction or post-inference interpretability, they get classified as ante-hoc methods. In cases of models that are not IIM, such as deep learning models or large ensembles, a post-hoc explanation framework needs to be applied after the fact, post-model learning that is separate from the model itself. Under the post-hoc category, we can define two subroups, those post-hoc methods that only work for specific data modalities or models, and model-agnostic explainability methods. The post-hoc approach attempts to address the trade-off between high interpretability and high predictive performance by providing methods to add-on interpretability to complex but high-performing models. Thus, instead of developing an IIM which, as we have seen, suffers under certain constraints either as part of the assumptions or model prediction capabilities, post-hoc explanations can be paired with a wide swath of complex AI models to extract interpretability without understanding the inherent reasoning of the model itself. 

There has been recent work on making deep neural models ante-hoc interpretable including approaches like self-explainable neural networks (SENNs) and CoFrNets. SENNs rely on staggered generalisation of linear models as a function of locally interpretable functions and representations of the input. The example of the model structure can be seen in \cite{alvarez2018towards}. So far, SENNs have only been tested on a few tabular and extracted imaging data. CoFrNets, on the other hand, rely on the mathematics of continued fractions (CFs), typically represented as a ladder-like sequence: $a_0+\frac{b_1}{a_1+\frac{b_2}{a_2+\cdots}}$ which can represent any real number and any analytic function \cite{churchill2014ebook}. Using the repeating structure and linear activations in between layers (which can be shown are sufficient to achieve universal approximation), the input is passed to each layer and a linear number of weights is learned for each layer instead of the quadratic amount in classic neural networks \cite{puri2021cofrnets}. Per-example feature importance is then computed using the gradients throughout the ladder as a function of the input and global attributions taking advantage of the ladder being a representation of a multivariate power series with the coefficients of the two forms mapping one-to-one. Further work is needed, however, to establish the high predictive performance compared to other black-box deep learning models.


Both of the previous methods touch upon conceptual learning, that is, when deep learning models learn from data, they, in fact, extract key concepts important for making the predictions. If one could extract these internal learning concepts, then one would be a lot closer to understanding the internal reasoning of the models, thus making them more explainable. Recently published work builds on this approach by appending a concept generation module (encoder) to jointly optimise for interpretable concepts on top of a latent encoder used for image classification. The concepts are then passed to a decoder to estimate reconstruction error and thus provide an indirect way to measure how "interpretable" the concepts are by how well they reconstruct the input. The concepts learned are individually informative by enforcing a fidelity loss. One does not have to rely on the concepts being directly supervisable (i.e. ground truth annotations are not available) because using self-supervision through incorporation of a separate loss function as an auxiliary task allows extension to more learning problems. Even though the framework adds components to existing deep learning models, most can be discarded after training and explanation-generating module can be used at prediction time \cite{sarkar2022framework}. More experiments need to be done to show good explainability and predictive performance on datasets with varying sizes, including those that are not just image-based, but the conceptual learning paradigm offers a motivating approach to ante-hoc explainability of black-box neural networks.


A more popular field of inquiry has been post-hoc methods which provide interpretability and limited explainability after model training, inference, and deployment. These include feature attribution (global and local) methods like Shapley values which adapt concepts from game theory to investigate feature importance and contributions to predictions. A Shapley value is calculated for each feature $j$ with feature value $x_j$, usually based only on the test set for risk prediction, by using sets of all possible unions with $m$ features except feature $j$. The value is the difference between the results of the characteristic function $v$ on $M$ (the set of all features) and $S$ (the subset of $M$ without feature $j$). The Shapley value is then averaged across the marginal contributions of all possible combinations of the feature unions:
$$
\varphi_j(v)=\sum_{S \subset M \backslash\{j\}} \frac{|S| !(m-|S|-1) !}{n !}\left(v\left(S \bigcup\left\{x_j\right\}\right)-v(S)\right)
$$
Shapley values are simultaneously useful for both finding the contribution of each feature to an individual prediction as well as providing global explainability across all samples \cite{lundberg2017unified, ibrahim2020explainable}. Calculating the value itself is usually computationally expensive even on small test datasets, at which point kernel sampling estimators are used to summarise the process. By defining a specific kernel in KernelSHAP allows for a smaller number of necessary samples by a combination of sampling and penalized linear regression. The method does not, however, take feature dependence into account, especially in its kernel version leading to over-weighing outlier data points. The tree version of SHAP addresses this but its reliance on conditional expected predictions is known to produce noise as non-intuitive feature importance values \cite{molnar2020interpretable}.

A slightly earlier proposed method, local interpretable model-agnostic explanations (LIME), focuses only on explaining local or individual predictions. LIME relies on introducing perturbations on the training data and observing how they affect the corresponding predictions given an interpretable model like some of the IIMs mentioned earlier all the while approximating the black-box model in the neighbourhood of the sample \cite{ribeiro2016should}. In applications to clinical risk prediction, the method has been found to produce unstable feature importance due to random perturbations. \cite{zafar2019dlime} propose DLIME (D standing for deterministic), which uses hierarchical clustering to group the training data together and kNN to select the relevant cluster of the new instance being explained instead of relying on randomness. Evaluated on clinical risk problems, DLIME presents a more stable version of instance-wise interpretability than its predecessor. 

LIME and SHAP present foundational interpretability methods due to their widespread adoption and relatively clear heuristics. Local Interpretable Visual Explanations (LIVE) build on the LIME paradigm by prioritising the visualisation of the interpretability results while keeping the random data perturbations approach and local point variation. While LIVE does not require an interpretable input space since it approximates the black box model directly from the data, it has no theoretical foundation like Shapley values. BreakDown uses a non-local approach to post-hoc, model agnostic interpretability that permutes the data samples based on relaxed loss effect constraints instead of relying on random sampling \cite{staniak2018explanations}. The space of features is searched to look for a set of variables where the model prediction will be close to the average expected value across all model predictions. To reduce the computational complexity of the search, a greedy approach is taken. These methods, however, decompose final prediction into additive attribution components and usually do not work well for models with significant inter-feature correlations.

Manipulating feature input and evaluating its effects on the outcome is a common set-up of many interpetability methods. The contrastive explanations method (CEM) finds what features should be minimally and sufficiently present to justify its classification and analogously what should be minimally and necessarily absent. This method was evaluated on deep learning models for applications including brain activity prediction but generating explanations of the form "an input $x$ is classified in class $y$ because features $f_j, \cdots, f_k$ are present and because features $f_m, \cdots, f_p$ are absent" \cite{dhurandhar2018explanations}. As compared to other methods like LIME, CEM highlights what features need to be present for a specific classification and not just list positively or negatively relevant features that may not be necessary or sufficient to justify the classification. Finding this set of features is defined as an optimisation problem where the modified data is sampled close to the original by using reconstructions from an autoencoder. These types of explanations might be more intuitive to many clinical applications especially in diagnosis as highlighted by \cite{miller2021contrastive}. Finding the contrastive explanations can be aided using an adversarial model, for example, but the method's limitations of defining what contrastive examples would represent in complex data cases like timeseries has not been sufficiently explored. An image pixel can be made into a constrastive case by being set to 0 but greyscale examples would encounter challenges while a similar approach for timeseries would not be robust \cite{luss2021leveraging}.

Post-hoc methods are generally model agnostic and apply to a myriad of black box models by directly manipulating the input either through finding a subset of data points or examples that are more "informative" or explainable to the prediction or a subset of the feature space that does the same. ProtoDash is a method that generated prototypes from the data and assigns non-negative weights to each of them depending on their importance to the prediction problem. Since each set of prototypes can be weighted, ProtoDash offers a robust response in cases of class imbalance which have been shown to affect the fidelity of earlier interpretability methods \cite{shao2021introducing}. Effectively, ProtoDash is an instance-based interpretability method as the prototypes it samples are examples of samples that would be more important for a prediction rather than features as such. Other earlier instance-level methods like K-medoids and MMD attempt to do the same but ProtoDash provides more general rule explanations with more robust kernel performance that help identify cases of low test accuracy while being relatively constant in its interpretability \cite{kim2016examples, nguyen2020quantitative}.

So far we have only discussed methods that manipulate the data space in some form to evaluate feature or example importance for prediction, but other post-hoc paradigms exist that more directly measure model learning. In deep learning, computed gradients are used to update weights through the learning procedure. In cases such as images or timeseries with spatio-temporal dimensionality, visualising the absolute weights of the gradient updates across those dimensions can indicate what part of the example image or timeseries is more influential for a prediction averaged across all examples \cite{itti1998model}. Examples include using visual saliency maps to highlight sections of ECG signal important for arrhythmia classification or x-ray images with lung cancer detection which can aid in learning while providing clinicians with more explainable model predictions \cite{jones2020improving, arun2021assessing}. These saliency maps are often sensitive to noise in the input and several approaches have been proposed to address these limitations including by adding noise to the data multiple times and then taking the average of the resulting saliency maps for each example like with SmoothGrad \cite{smilkov2017smoothgrad}. Newer methods like saliency guided training have features iteratively masked with low gradient values and then minimise a combined loss function of the KL divergence between model outputs from the original and masked inputs, and the model's own loss function. Experiments have shown the latter maintains higher predictive performance while simultaneously reducing noise sensitivity \cite{ismail2021improving}. General limitations of saliency maps in clinical risk prediction have been their tendency to represent clinically non-useful information as important as evaluated by physicians and simply localising the area of the example does not add further intepretability on why that area is important for model learning which can appear misleading to domain experts \cite{adebayo2018sanity}. They have also been vulnerable to adversarial attacks when explanations remain unchanged even after significant adversarial perturbation in the input has changed model behaviour \cite{gu2019saliency}. 

Visualising weights from the learning process to highlight which aspects of the data the model focuses on for training is the underlying heuristic of attention mechanisms which became a highly popular deep learning model-specific method to achieve better predictive performance while providing more interpretable results. An attention function maps a query and a set of key-value pairs where the output is computed as a weighted sum of the values and the weights are computed by a compatibility function of the query with the corresponding key. Multi-head attention parallelises the attention layers to project multiple mappings simultaneously. A comparison of the two frameworks can be seen in \cite{vaswani2017attention}. Compared to recurrent layers for learning sequences of data requiring $O(n)$ operations, a self-attention layer connects all positions with a constant number of sequentially executed operations, thus being computationally cheaper. In the example of timeseries data, the first step is to initialise the attention vectors mapped to the features such that for each feature $j$ an attention vector $\boldsymbol{a}_j$ of length $T$ (number of time-steps) is learned with $\left|\boldsymbol{a}_j\right|=1$:
$$
\boldsymbol{X}_{\text {new }}=\boldsymbol{A} \odot \boldsymbol{X}
$$
where $\boldsymbol{X}$ represents the $T \times m$ input data ($m$ number of variables). Once sofmax normalisation is applied, $\boldsymbol{a}_{j t}$ can be interpreted as a contribution of feature $j$ within a fixed time step $t$. \cite{gandin2021interpretability} extend this for global interpretability by applying the softmax to the transposed input with the same notation:
$$
\boldsymbol{a}_t=\operatorname{softmax}\left(\boldsymbol{x}_t \boldsymbol{W}_t\right)
$$
By aggregating values $\boldsymbol{a}_{j 1}, \ldots, \boldsymbol{a}_{j T}$ through time one can get the global contribution of the $j$-th feature $\boldsymbol{x}_t$ and weights $\boldsymbol{W}_t$ at time $t$. Not only is an absolute ranking of features provided, but one can use the softmax activation matrix to extract patient-level feature importance as well as visualise the attention weights using heatmaps (attention maps) both in timeseries and imaging problems, similar to the saliency maps scenario despite early criticism of the lack of explainability of such maps for standard attention mechanisms \cite{jain2019attention}. Some results do indicate better interpretability of the attention plots but further experiments are still needed to evaluate the fidelity of this approach in achieving the proposed changes in interpretability.

Since being originally proposed in 2017 as an alternative to convolutional and recurrent behaviour in deep learning pattern recognition, transformers (models based solely on attention mechanisms) have become robust prediction models in various applications for clinical risk prediction, including ICU outcome prediction, ECG signal diagnosis, and blood pressure response \cite{vaswani2017attention, xu2022time, kamal2020interpretable, zhang2019application, girkar2019predicting}. Some limitations include inability to distinguish between positive and negative associations among features and output, lack of propagated relevancy through the attention layers partially addressed through a new layer-propagation strategy, and unstable interpretability with covariate shift \cite{chefer2021transformer, baric2021benchmarking}. 


We have seen in this section a combination of methods that try to address the interpretability aspects of explainability using both innovative ante-hoc glass-box approaches to deep learning as well as what have become off-the-shelf methods like LIME and SHAP. Most of the methods share limitations on their robustness to random noise, permutation and covariate shift, as well as a lack of clinical evaluation of their inherent interpretability qualities.

\subsubsection{Global vs. Local}
Another axis of separation for explainability methods is on the global (feature-space) versus local (sample-space) levels. A local or instance-based (used synonymously for some methods) explanation deals with explaining the prediction of an individual sample, whereas a global explanation attempts to provide a summarised view of the entire model. In the clinical risk prediction realm, both can be valuable despite the latter being more popular in recent years of research. Local explanations can be used for explaining individual patient trajectories and predictions all the way to their grouping in phenotype clusters, for example, while global explanations can give one a more holistic view of the impacts of certain risk factors on a population-size patient cohort \cite{elshawi2021interpretability}. Local interpretability methods usually provide more stable and linear results for explanations which are partially why methods in deep learning like guided backpropagation, Grad-CAM, and SmoothGrad saliency maps (described in the gradient-based section later on) have remained popular over time.

From the interpretability methods covered earlier, LIME and SHAP also support local explanations by providing a result for each instance in the sample space. The instance is then perturbed according to different criteria in the case of LIME while SHAP uses a more robust generation of individual explanations because of its guarantee of a fair distribution effect among the features coming from a more established theoretical background for the latter method. SHAP provides high efficiency of feature values computation especially in the presence of multi-collinearity among the features while providing a theoretical framework to prove, under certain assumptions, the unique existence of such local models \cite{lundberg2017unified}. An extension of LIME which resolves the significant dependence on weights assigned to different perturbed samples is ILIME (I standing for influence-based). ILIME uses the influence of perturbed instances on the instance of interest as well as their Euclidean distance to estimate the weights with influence functions \cite{elshawi2019ilime}.

ProtoDash produces as an explanation representative instances or prototypes from the data that estimate the underlying distribution. As mentioned earlier, ProtoDash generates these samples by assigning non-negative weights of importance. The generation depends on the property of sub-modularity of its scoring function to find samples efficiently. The computational cost of such guarantees is significant, thus a class of approximate sub-modular functions are used \cite{gurumoorthy2017protodash}. 

MAPLE or model agnostic supervised local explanations method uses random forests to achieve neighbourhood selection in local linear modelling. The neighbourhood for each instance is defined by each instance's occurrence in the same leaf node as the instance to be explained. A score is then assigned to each feature based on its frequency of splitting a node at the root of the trees in the forest. The set of most important features is then used for approximating the linear regression coefficients in the final explanations. The great advantage of MAPLE is that it uses random forests which suggests better predictive performance while still being interpretable \cite{elshawi2021interpretability, plumb2018model}. 

Counterfactual explanations mentioned earlier in the introduction as a relatively recent proposal for a more general and abstract way to explain models while addressing fairness and bias concerns also can be classified as a local interpretability method. In short, "a counterfactual explanation of a prediction describes the smallest change to the feature values that changes the prediction to a predefined output" as explained by \cite{molnar2020interpretable}. Creating these explanations for machine learning models has been first proposed by \cite{wachter2017counterfactual} who suggest minimising the loss:
$$
L\left(\boldsymbol{x}, \boldsymbol{x}^{\prime}, y^{\prime}, \lambda\right)=\lambda \cdot\left(\hat{f}\left(\boldsymbol{x}^{\prime}\right)-y^{\prime}\right)^2+d\left(\boldsymbol{x}, \boldsymbol{x}^{\prime}\right)
$$
The loss consists of the quadratic distance between the model prediction for the counterfactual $x^{\prime}$ and the desired outcome $y^{\prime}$ and the distance between the sample $x$ to be explained and the counterfactual $x'$. $\lambda$ is a parameter over which one maximises by iteratively solving for $x^{\prime}$ and
increasing $\lambda$ until a sufficiently close solution is found. The distance function $d$ is the Manhattan distance weighted with the inverse median absolute deviation (MAD) of each feature $j$ in feature set $M$.
$$
d\left(\boldsymbol{x}, \boldsymbol{x}^{\prime}\right)=\sum_{j=1}^M \frac{\left|\boldsymbol{x}_j-\boldsymbol{x}_j^{\prime}\right|}{M A D_j}
$$
The total distance is the sum of all absolute differences of feature values between sample $\boldsymbol{x}$ and counterfactual $\boldsymbol{x}'$. The usual advantage of Manhattan distance over Euclidean for robustness to outliers stands as usual. 



One of the limitations of counterfactuals generally is their lack of uniqueness, i.e. multiple counterfactual explanations can be provided for the same prediction. The method above cannot efficiently handle categorical features with many different levels since it depends on running the method separately for each combination of feature values. An approach was proposed by \cite{back2020parallel} using four-objective loss that allows for multi-category counterfactual optimisation with a nondominated sorting genetic algorithm. Extended work on applications in healthcare like for glaucoma detection has shown better explainability than classical methods like saliency maps \cite{chang2021explaining}.

K-nearest neighbours (kNN) is a popular distance-based machine learning method which assigns the class of each point based on its k closest neighbours. Because it is a nonparametric model using all of the available data to make a prediction for each sample, it is in-effect an instance-based interpretable machine learning model. The lack of learned parameters implies no global interpretability as such but using the k neighbours that were used for the prediction for each sample, one can provide limited interpretability for that prediction. This becomes true in cases of constrained feature space where heuristic or simple analysis of neighbouring points' features is manageable but not in cases where a large number of features is used \cite{molnar2020interpretable}. This capability was in an indirect way used in the methodology for DLIME when the 'important' and 'explainable' clusters are identified using kNN in combination with hierarchical clustering first used to generate an intermediate clustering stage \cite{zafar2019dlime}. Further work on integrating local explainability propensity of kNN methods in combination with other methods like, for example, clustering Shapley values in local regions around a specific sample have not been sufficiently studied. They might provide an avenue in addressing instability in local explanations due to random perturbations as was done in the case of DLIME. 

Other methods mentioned earlier like ProfWeight, Boolean rules, CoFrNets, and others all provide, as of now, global explainability exclusively which means they provide feature importance and explanations across the entire sample set without resolution to the sample level. One can then say that such methods provide an average estimate explanation for the model and the data and are useful in understanding general behaviours of the system. 

\subsubsection{Model-agnostic vs. Model-specific}
The division of explainability methods also includes specifications of the levels of model nature the methods can apply to. Some methods are model-agnostic, meaning that they can be used to provide interpretable results for a variety of models. This is the case for Shapley values, for example, which can be used to provide interpretability for extreme gradient boosted decision trees, deep learning models, as well as some clustering algorithms \cite{brandsaeter2022shapley}. A more general method is permutation importance because this method relies on manipulating the input feature sets rather than having anything to do with explaining the internal model dynamics, so it can be applied across different machine learning models as long as a relatively distinct subgroup can be formed for the feature space \cite{molnar2023model}. An obvious problem is the computational cost explosion resulting in cases of higher dimensionality.

A subgroup of model-agnostic methods can be described as visualisation methods because of their focus on creating curves and plots describing the output's associations to input features after prediction. Examples like partial dependence plots (PDPs) rely on using a partial dependence function of the output with respect to subsets of the feature space. Values in the input space are varied slightly with respect to their marginal distribution with the partial dependence function approximated by a statistical model from which plots of associations between covariates and output are created \cite{friedman2001greedy}. The assumption is that because this is done for a single predictor at a time, that predictor is relatively independent of the remaining set of features on average which might not stand. This limitations combined with PDP's weakness of dealing with extrapolations in the feature space have led to the introduction of individual conditional expectation (ICE) plots. In ICE, instead of plotting the feature average partial effect on the output, one plots the estimated conditional expectation curves, each reflecting the output as a function of the feature effectively disaggregating divergent effects. The PDP curve in the plots is then simply the average of N ICE curves behaving as a local interpretability model that plots a curve for each instance of the data \cite{agarwal2020interpretable}. Some limitations of ICE plots include their overcrowding due to curves created for each sample instance in the plot, difficulty detecting outliers and overfitting extrapolations from view. Furthermore, the complexity of a higher feature space makes this method practically infeasible. Some recent advances like ICE feature impacts address these by allowing comparisons between different ICE plots and defining a metric, takeing into account all samples as well as outliers over which feature contributions are averaged to identify the most impactful features (not the same as PDP where the curves and expectations are themselves averaged for each covariate contribution), thus severely decluterring the plots \cite{yeh2022bringing, goldstein2015peeking}.

In some cases there is additional flexibility in the types of explanations generated as well. Instead of getting just feature importance rankings or plots, some model-agnostic methods can provide other types of explanations in parallel like linear formulations or graphic interfaces \cite{ribeiro2016model}. The great benefit of this approach is the ability to use a variety of complex, high performance, and uninterpretable models which can be rendered more interpretable without having to make sacrifices on performance by choosing simpler and more transparent models. 

Model-specific methods tend to only work on a certain subset of machine learning models and methods. Cases include saliency maps developed for elucidating image regional weighing in the learning process of deep learning models similar to attention maps. Another example described earlier includes generalised linear rule models (GLRMs) which learn a linear combination of conjunctions using link functions like logit in the case of logistic regression but which only work on this set of models respectively \cite{arya2019one}. Depending on the intended usage, these two approaches offer different advantages. If the user in clinical risk prediction wants to focus on more predictive and high-performing models, then model-agnostic methods often can aid in final interpretability design if existing methods have not been proposed and sufficiently evaluated for that model before. On the other hand, some might take the route of choosing a simpler yet more explainable model which they can couple with a model-specific method to provide more clinical insight and explainability all the while giving satisfactory predictive performance. In fact, it is possible to take the approach of using different interpretability methods and evaluating their comparative explainability and performance \cite{meng2021mimic}. There is no one size fits for this type of explainability challenge and the different contexts for each problem should be investigated robustly before the implementation stage. 

\subsubsection{Perturbation- vs. Gradient- vs. Instance-based}
Some, albeit rare, classifications of explainability methods include their grouping based on how the predictions and associated interpretability results are calculated with respect to the pairing of input and output. Such groupings include gradient- and perturbation-based (sometimes confusingly called instance-based) approaches. Perturbation-based methods perturb the model around the prediction to infer the feature importance of the input-output pairing \cite{alvarez2018robustness, alvarez2017causal}. Perturbations of images, for example, are performed by removing or inserting pixel or patch values to generate saliency maps at different levels of occlusion. Pixel-wise perturbations tend to be more spatially discrete and represent saliency more accurately in terms of location, yet their higher granularity leads to worse representation of the semantics of salient objects \cite{ivanovs2021perturbation}. Finding the right set of perturbations to apply remains a challenge across all cases.

An example of the gradient-based approach is integrated gradients that can help visualise input importance by calculating the integral of the gradients of the output prediction with respect to the input image pixels without any modification to the original network. Integrated gradients are defined as the path integral of the gradients along the straight line path from the baseline $\boldsymbol{x}^{\prime}$ to the input $\boldsymbol{x}$. The integrated gradient along the $j^{t h}$ dimension (can be understood as $j^{t h}$ feature) for an input $\boldsymbol{x}$ and baseline $\boldsymbol{x}^{\prime}$ with a straight line slope of $\alpha$ is defined as follows:

\begin{equation}
\operatorname{IG}_j(\boldsymbol{x})::=\left(\boldsymbol{x}_j-\boldsymbol{x}_j^{\prime}\right) \times \int_{\alpha=0}^1 \frac{\partial F\left(\boldsymbol{x}^{\prime}+\alpha \left(\boldsymbol{x}-\boldsymbol{x}^{\prime}\right)\right)}{\partial \boldsymbol{x}_j} d \alpha
\end{equation}

Another sub-domain of gradient-based methods are saliency-based methods which use gradient values either raw or normalised to infer salient features like in saliency maps described earlier. \cite{alvarez2018robustness} highlight the weakness of such methods under robustness checks despite their advantages like simple formulations and lack of reliance on model pliability. Saliency or heatmaps have also been criticised for insufficient correlation with the network which they are meant to interpret thereby undermining their reliability as interpreters of the model. To address this concern, \cite{qi2019visualizing} propose Integrated-Gradients Optimized Saliency (I-GOS), a method which optimises a heatmap with the objective of maximally decreasing classification scores on the masked image. This is done by computing descent directions based on integrated gradients thus avoiding local optima and speeding up convergence. Gradient-based methods are often, however, not invariant under simple transformations of the input, and are very sensitive to the choice of reference point. 

Further well-known examples of gradient-based methods include Deep Learning Important FeaTures (DeepLIFT) and deconvolution approaches \cite{shrikumar2017learning}. Deconvolution approaches like the deconvolutional network (Deconvnet) rely on deconvolving the network, going from neuron activations in the given layer back to the input. The reconstruction highlights part of the input that is most strongly activating the neuron \cite{springenberg2014striving}. Due to the zero-ing out of negative gradients, both approaches tend to not capture inputs that negatively contribute to the output prediction \cite{shrikumar2017learning}. 

Layer-wise Relevance Propagation (LRP) is another attribution method that propagates the prediction backwards in the neural network and compared to its perturbation- and gradient-based competitors it does not involve multiple computationally expensive neural network evaluations. It instead utilises the graph structure of the deep neural network to generate explanations \cite{montavon2019layer}. The relevance score that is propagated is estimated for neurons $j$ and $k$, for example, based on those above them: 
\begin{equation}
{R}_j=\sum_k \frac{z_{j k}}{\sum_j z_{j k}} {R}_k
\end{equation}
The quantity $z_{j k}$ (neuron activation) models the extent to which neuron $j$ has contributed to make neuron $k$ relevant (denominator guaranteeing that the relevance is propagated downwards equally). The last layer relevance score is the pre-activation value corresponding to the class for which relevance scores are desired, and are obtained from the forward pass of an input. The propagated values to each layer are conserved and their sum is constant. LRP has been applied in many settings to offer interpretability to deep learning models which will be highlighted later under the applications section. LRP has been shown to give better relevance or attributions than those by DeepLIFT without relying on the need for a reference input \cite{grezmak2019interpretable}. LRP suffers from its dependence on ReLU activations resulting in non-negative attribution maps thereby limiting the interpretability of the application, later addressed by an extension of the transformer using LRP-type of relevance score to compute the relative relevance of heads in the attention mechanism using GELU instead. Conservation of the relevance score in the self-attention is then maintained through a normalisation step of the scores at the layers of the model. Class-specific interpretability can be obtained by extensions like Contrastive-LRP (CLRP) and Softmax-Gradient- LRP (SGLRP) where the results of the class to be visualized are contrasted with the results of all other classes, to emphasize the differences and produce a class-dependent heatmap \cite{chefer2021transformer}.

Other methods rely on manipulating the layers of the neural network itself similar to attention layers intuitively. Class Activation Mapping (CAM) effectively removes the last fully connected layers and replaces the MaxPooling layer with a global average pooling (GAP) layer after which the weighted average of features is extracted to form a per-class activation map (overlayed on the input like a heatmap) \cite{zhou2016learning}. This makes it a non-gradient-based method but extensions of it like Gradient-weighted CAM (Grad-CAM) and Grad-CAM++ which propagate gradients through the network in obtaining the final heatmaps for input regional relevance. Other alterations of the method include methods like Eigen-CAM which extracts the principle components from the convolutional layers and helps against classification errors made by fully connected layers in CNNs without needing to backpropagate gradients or use feature weighting. Early experiments have shown promising results albeit further evaluation on different datasets is needed alongside clearer analysis of the relative interpretability compared to other mask methods \cite{vinogradova2020towards, muhammad2020eigen}. 

Both perturbation- and gradient-based methods can be described as attribution-based as they establish a link of attribution between some area or form of the input to the output prediction. A disadvantage of the perturbation-based methods is the large number of possible combinations if one blindly attempted at going through all possible ways of perturbing the input without relying on approximations. Additionally, for different samples in the same dataset belonging to the same class contradicting explanations can be generated resulting in decreased user trust. In a way, gradient-based approach address this by using the gradient as a proxy for these changes. relying on evaluating numerous feature subsets or solving an optimiza- tion problem for each instance of data. While gradient-based methods provide faster explanations, they also tend to be less accurate \cite{adebayo2018sanity}. An advantage of perturbation-based methods, on the other hand, is their ability to query models repeatedly and their model-agnostic nature \cite{ivanovs2021perturbation}. 

Another group of instance-based methods seeks to remedy the disadvantages of perturbation- and gradient-based methods by optimising for the fidelity of instance-level explanations. An example of such methods are amortized explanation methods (AEMs) which generate explanations by learning a global selector model that efficiently extracts locally important features in an instance of data with a single forward pass using an objective that measures the fidelity of the explanations. AEMs assess these selections with a predictor model for the output. L2X and INVASE fit the predictor and selector models jointly, often referred to as joint amortized explanation methods (JAMs) \cite{jethani2021have}. Since they are instance-based methods, JAMs have been previously deployed to explain mortality predictions to a patient-level \cite{yoon2018invase}. Some problems with JAMs include their propensity to encode predictions with the learned selector and failure to select features involved in control flow (features involved only in branching decisions/nodes of tree structured generative process). An example of this could be mortality predictions for patients with chest pain using EHRs as \cite{jethani2021have} point out. Blood troponin levels can be a control flow feature where abnormal values indicate that cardiac imagining should be used to assess severity whereas normal values indicate that the chest pain may be non-cardiac. In this case, using JAMs to interpret the prediction will not represent the true role of troponin in patient mortality. To address these issues, \cite{jethani2021have} propose REAL-X and EVAL-X, a framework that combines efficient instance-level explanations with a single forward pass while also detecting when predictions are encoded in explanations without making out-of-distribution queries using approximations of the true data generating distribution. Other examples of instance-based methods include ProtoDash (and earlier iterations like influential instances and MMD-Critic) and counterfactuals described in detail in earlier sections. This section has included a myriad of different methods which have often developed as a response to weaknesses present in each other. There is no perfect explaianbility method as we still also lack a robust explainability metric to reliably compare them by.

The section was meant to highlight diverging methodological pathways in approaching the same problem of using perturbations of the input or gradients of complex deep learning models in extracting some form of discernible explanations from machine learning problems. Attempts to do so in clinical risk prediction will be described in the next sections.

\subsubsection{Concept-based}

Compared to previously described methodologies, concept-based explainability relies on using internal conceptual learning of the models in extracting semantically-meaningful latent variables or concepts. Work-based on conceptual learning with concept activation vectors (CAVs) first and foremost seeks to learn a latent space that would correspond more closely to human-level insight. Sometimes, these concepts are obtained through PCA dimensionality reduction (PCANet) or by using representative training patches to explain a prediction or even by using human-labelled concepts in the estimation of concept scores \cite{chan2015pcanet, chorowski2019unsupervised, chen2019looks, koh2020concept}. Conditional VAE models have also been proposed to model the causal effect between concepts and output predictions \cite{goyal2019explaining}. Testing with CAVs (T-CAV) uses directional derivatives to measure the 'conceptual sensitivity' to a high-level concept as learned by a CAV i.e. the sensitivity of the output to changes in the input towards the direction of a concept at activation. The score (per concept per class) is the fraction of inputs whose l-layer activation vector was positively influenced by the concept (the magnitude of this influence is not taken into account). A major limitation of this approach is relying on, possibly biased, human-labeled examples of that concepts and the space of possible concepts is not necessarily limited a priori. 

Automated Concept-based Explanation (ACE) is meant to address some of these limitations by automating the selection of meaningful concepts. A concept as such can be described as meaningful, coherent, and important \cite{ghorbani2019towards}.  Work on more diverse data is still to be evaluated but a promising automated framework for conceptual explainability makes great inroads in alternative interpretability of opaque large deep learning models. 


An example of a concept-based interpretability method that does not require generative modelling, retraining, or reliance on clustering some latent space of a large deep learning model, is the Shapley value extension ConceptSHAP. In ConceptSHAP, subsets of concepts are selected based on completeness as measured by a concept's score to reconstruct the output. The completeness score is then just the average accuracy of predicting the label using the concept scores done by some MLP. In other words, measuring the accuracy achieved by the concept score means measuring how “complete” the concepts are without relying on a score implicitly assuming a first-order relationship between the concepts and the outputs like ones does with T-CAV and ACE. These methods also make no assurances of the completeness of the concepts in explaining the model such as how many are needed for sufficient interpretability and their saliency scores may fail to capture concepts that have non-linear effects. Both of these limitations are addressed by ConceptSHAP where concepts are extracted in an unsupervised manner without these assumptions and where relevance is determined by Shapley values. However, a small MLP is involved in extracting concept embeddings, which makes the method not fully interpretable and self-contained \cite{kamakshi2021pace, yeh2020completeness}.  

The role of causality has been mentioned earlier in implementing some interpretability methods for machine learning models. An entirely separate review can be written for causality and causal machine learning's role in explainability research but here we will present a method intersecting the former with concept-based explanations called Causal Concept Effect (CaCE). CaCE explores the causal effects of the presence or absence of a human-interpretable concept on the model prediction. This method as well as some of the previous ones are global i.e. providing average interpretability across all instances of the data provided and as such are more vulnerable to confounding of concepts. Investigating the presence or absence of concepts relies on using conditional generative models (VAEs for example) to generate counterfactuals and approximate the causal effect of concept explanations. For each sample in the test set, the difference in the model outputs is calculated for the original sample and for the counterfactual as generated. CaCE, by definition, is the average of these differences. T-CAV as we have mentioned remains vulnerable to collinearities between concepts and class labels whereas CaCE learns to measure the causal effects of the explanations. In some cases, CaCE can be computed directly if one has access to the generation process of the data (ground truth CaCE). One limitation of this method is that it is rather a metric for estimating the causal effects of pre-determined concepts rather than identfying the same for added explainability. As such, it would be best combined with other explainability methods as an added metric for concept-based expanation evaluations. Furthermore, relying on the data generating process to perturb concepts is unsustainable both because of the myriad of combinations, biases, and costs associated as well as the vulnerability to the weaknesses of the generative models used. Further work needs to be done in exploring the relationship between causal effects, causality, explainability, and conceptual learning for more human-friendly yet automated ways to provide explanations. 

Concept-based explanations remain a source of insufficiently explored research especially compared to other areas of explainability. Moving away from image-based examples, concept-based methods have also been applied to text with INLP, CausaLM, and S-Learner \cite{ravfogel2020null, kunzel2019metalearners}. Each makes different assumptions on the model structure and their approach to estimating the causal effects of changes in the concept level. In CausaLM, for example, causal model explanations are generated using counterfactual language representation models based on fine-tuning of deep contextualized embedding models (like BERT) with auxiliary adversarial tasks derived from a causal graph \cite{feder2021causalm}. Only recently has there been work on proposing a real-world dataset for benchmarking approaches and metrics to compare concept-based explanation methods which shows how limited current approaches are \cite{abraham2022cebab}. Their approximate counterfactual baseline outperforms all earlier mentioned methods at capturing both the direction and magnitude of causal effects showing just how far the field has to go to create robust concept-based explainability methods.

\subsection{Data Modalities}
\subsubsection{Imaging}
With the rise of deep learning models like convolutional neural networks (CNNs) in achieving success for image-based learning in the later half of the 2010s, applications to medical imaging seemed like the next step. As early as 2017, the first Food and Drug Administration (FDA) approved the application of AI in medicine and saw success with \textit{Arterys} for cardiac magnetic resonance images and since expanding to include liver and lung imaging, chest and musculoskeletal x-ray images, and non-contrast head CT images \cite{kaul2020history}. Within different sub-fields and specialisations in medicine, there exist different potential sources of image data including colonoscopies and the EUS platform in diagnosing malignant colon polyps or pancreatic cancer \cite{cazacu2019artificial}. As such, a large component of AI in medical imaging applications includes (early) diagnosis rather than risk prediction whether that be detection, analysis, image denoising in the case of MRI, generating clinical text description, or even just segmentation and visualisations \cite{renard2020variability, li2021assessing, monshi2020deep, alzubaidi2021novel}. Thus, a review of methods for clinical risk prediction using image data might be more sparse but the fact remains that a significant discussion of explainability in health care on image data cannot avoid mentioning these important applications as they contribute significantly to recent discussions of XAI in healthcare. 

Although deep learning models have continued to achieve high levels of accuracy in diagnosing conditions from medical images, concerns remain regarding their sensitivities as well as lack of explainability for clinical relevance. Since this data modality is more closely tied to opaque and black-box deep learning models as compared to the other data modalities, the medical and clinical AI community has had to discuss the specific needs for a meticulous assessment of these models \cite{liu2019comparison}. As image data is more easily analysed by humans in its original structure like in the case of text data, the patterns learned by AI might not be readily amenable to human identification as, for example, a conventional radiographic analysis would be \cite{lee2020machine}. This further highlights the importance in having explainabilty as an integral component of medical imaging AI applications. 

\begin{figure}[h]
  \centering
  \includegraphics[width=0.6\linewidth]{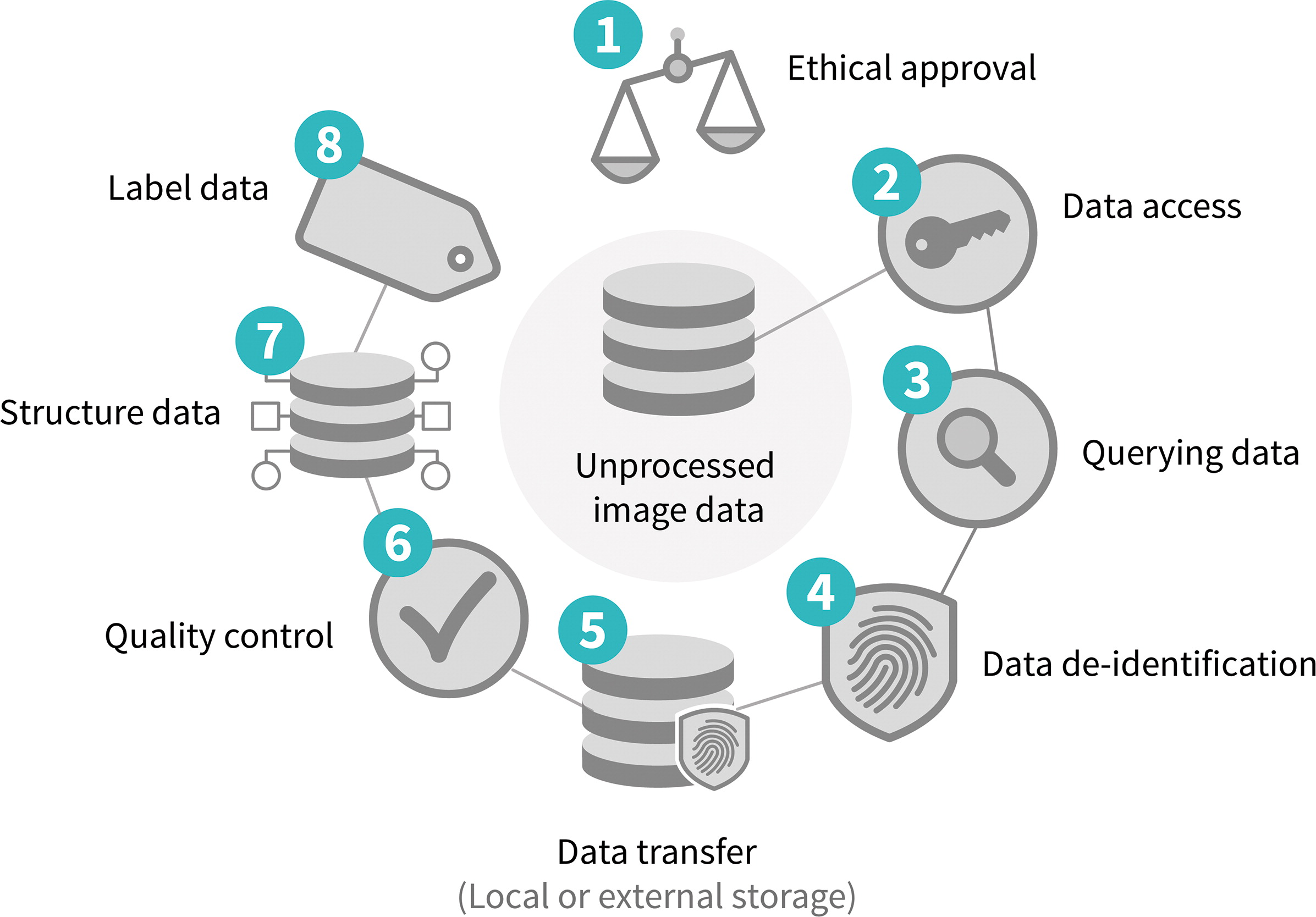}
  \caption{Diagram showing the data handling process for medical image data but which can be extended for other modalities as well \cite{willemink2020preparing}.}
  \label{fig:Medical Image Diagram}
\end{figure}

The structure of medical images is represented by a numeric matrix containing values of either 0 or 1 in the case of greyscale, or between 0 and 255, in the case of a colour image input. A key limitation remains also the need for extensive labelling, an expensive and slow process usually limited by human and time constraints, resulting in data augmentation techniques being even more important in increasing the sample size with the same amount of available labels \cite{chlap2021review, lundervold2019overview}. Figure \ref{fig:Medical Image Diagram} highlights steps in the usual order of image data handling. Another important consideration is that in most health care systems there is not widespread sharing of these patient data. Medical image data is often stored and analysed separately from other related patient data like electronic health records or genomics as well as data from different sites being in disparate silos, which is not optimal for research. When addressing the generalisability and trustworthiness of algorithms developed on these datasets, implementing limited explainability does not go far enough and being able to externally and clinically validate the models through multi-centre and interdisciplinary collaborations is a necessity \cite{willemink2020preparing}. Explainability in combination with multi-centre validation is, thus, crucial to achieving high-impact clinically meaningful AI algorithms. 

In some cases, other data modalities can be represented as images to take advantage of the deep learning models specifically designed for this data type. In cases of electrocardiogram (ECG) or other signal analysis, occassionally during the data processing pipeline the signal will be converted to a spectral image which is then given to the deep learning model as input \cite{murat2020application}. This, of course, means that the explainability analysis of such input remains limited with respect to the original input as the only image can be explained either in frequency or spectral domain and not the raw signal itself. But considering the noise and complexity of the original raw signal this can be seen as an improvement in creating human-understandable explanations for the predictions. In the following sections, we will detail approaches in AI for clinical risk prediction on medical images and later investigate the same with multi-modality in question, especially when combined with textual data from clinical notes. 

\subsubsection{Text} 
Following the previous discussion on medical image data, some images like radiology slides often come with associated clinical notes and text. The application of AI to clinical text is multi-faceted as it can apply to analysing a physician's notes on diagnosis and tests and extracting relevant patient characteristics, using medical image annotations by specialists for labelling or phenotyping, and even using chatbots, survey and feedback forms as a source for estimating patient well-being \cite{casey2021systematic, tanwar2022unsupervised, goh2021artificial, lin2021development}. Besides radiology reports, other clinical annotations of image data are also possible like electroencephalography reports when studying epilepsy or echocardiography reports in cardiovascular medicine \cite{casteleiro2020semantic, vaid2022using}. For some health conditions like sepsis, early symptoms might be difficult to identify so having access to previous clinical notes describing the patient state is especially useful in clinical AI risk modelling \cite{goh2021artificial, wu2021artificial}. NLP is often used to extract diagnostic information either from ICD coding schemes or clinical notes directly through, for example, discharge summaries \cite{li2020icd, sammani2021automatic}. One often needs to, however, analyse extensive amounts of clinical free text to extract relevant phrases (which can be a challenge to identify) for diagnostic purposes \cite{spasic2020clinical}. The variety of such NLP applications, thus, comes with sometimes unique and sometimes shared challenges specific to this data modality for clinical AI applications. 

And similarly again to medical image data, a large bottleneck in these applications is the annotations \cite{lybarger2021annotating}. Ideally, like in the previous modality case, unsupervised or semi-supervised learning paradigms can offer more robust answers to useful utilisation of clinical text for patient diagnosis \cite{spasic2020clinical, casey2021systematic}. 

Generalisability and external multi-centre validation is also potential solution to clinical text analysis when there are concerns about data representativeness between different health centres and their clinical text data. Most studies are limited by using only their local hospital or medical centre data which has been shown to lead to biased, overfitted, and disparate model solutions in clinical NLP \cite{spasic2020clinical, zhan2021structuring}. The value of openly available and relatively large datasets like Medical Information Mart for Intensive Care (MIMIC) is, thus, high allowing them to be used for standardised testing of models albeit relying on only a few of the same datasets without updates for developing and testing models over the years might eventually lead to similar problems. Using multiple centres for external validation of models in the clinical text which can differ greatly between health centres should be the optimal avenue to pursue \cite{kersloot2020natural}. Even in cases when data is available from different centres, standardising and preprocessing it consistently with large differences present still remains a challenge. 

The models often developed for text applications are often commonly shared with time-series implementations as well which we will discuss further in the electronic health records (EHR) section. Suffice to say that compared to modelling based on static or image data, the sequential pattern of text data requires special attention and the underlying model paradigm is to develop dynamic learning frameworks capable of internal updates as the model learns to "read" the sequence of text. Traditional rule-based and count-based models also exist which rely on the numbers and distributions of specific phrases in the text but they are obviously limited in their learning capacity compared to more flexible deep learning approaches \cite{casey2021systematic, minaee2021deep}. Large parallelised and parametrised approaches built on the transformer backbone have received increased attention in text generation and prompt responses with the generative pre-trained transformer (GPT) models and their variations. Pre-training such large models and applying them in clinical setting includes cases like mental health support but the inability of such models to adjust for tone, context, and body language of the situation remains an important limitation in clinical integration \cite{korngiebel2021considering}. Another major limitation is semantic repetitiveness, incoherence in long conversations, and internal contradictions \cite{elkins2020can}. As of now, using these relatively opaque models in healthcare with high-stakes interactions and emergencies comes with serious considerations that despite the overall popularity of the models in the general public leaves much to be desired in healthcare applications.

\subsubsection{Genomics} 
Genomic data consists of a sequence of nucleotides with the respective labels being A, T, C, or G belonging to different bases present in human DNA. As such, the sequence of these letters in a description of the genome or its specific parts, often called genes, can be understood almost as text data. To this end, methods using sequential modules like recurrent units in NLP and time-series analysis have been applied in detecting single nucleotide mutations (called SNPs) for the purposes of diagnosing so-called Mendelian diseases from a person's genomic profile \cite{alharbi2022review}. Studies have explored detecting variants causing complex eye diseases or early onset Alzheimer's disease often using either whole genome sequences or combining genomic data with other modalities like electronic health records and clinical notes and annotations  \cite{wang2022single, jo2022deep, venugopalan2021multimodal}. Genomic data represented in matrix format can be also viewed as image data. DeepVariant, a deep learning model, detects single-nucleotide variants and Indels from sequences given a fixed reference sequence \cite{kumaran2019performance, alharbi2022review}. DeepVariant relies on the dissimilarities in input images, in this case, the input and reference sequence, to perform the (image) classification task for genetic variant calling. In either case, deep learning methods like recurrent and convolutional and graph units (and their combinations) have become the standard implementation for these problems thus carrying their opaque and unexplainable nature with them to another data modality realm. 

\begin{figure}[h]
  \centering
  \includegraphics[width=0.8\linewidth]{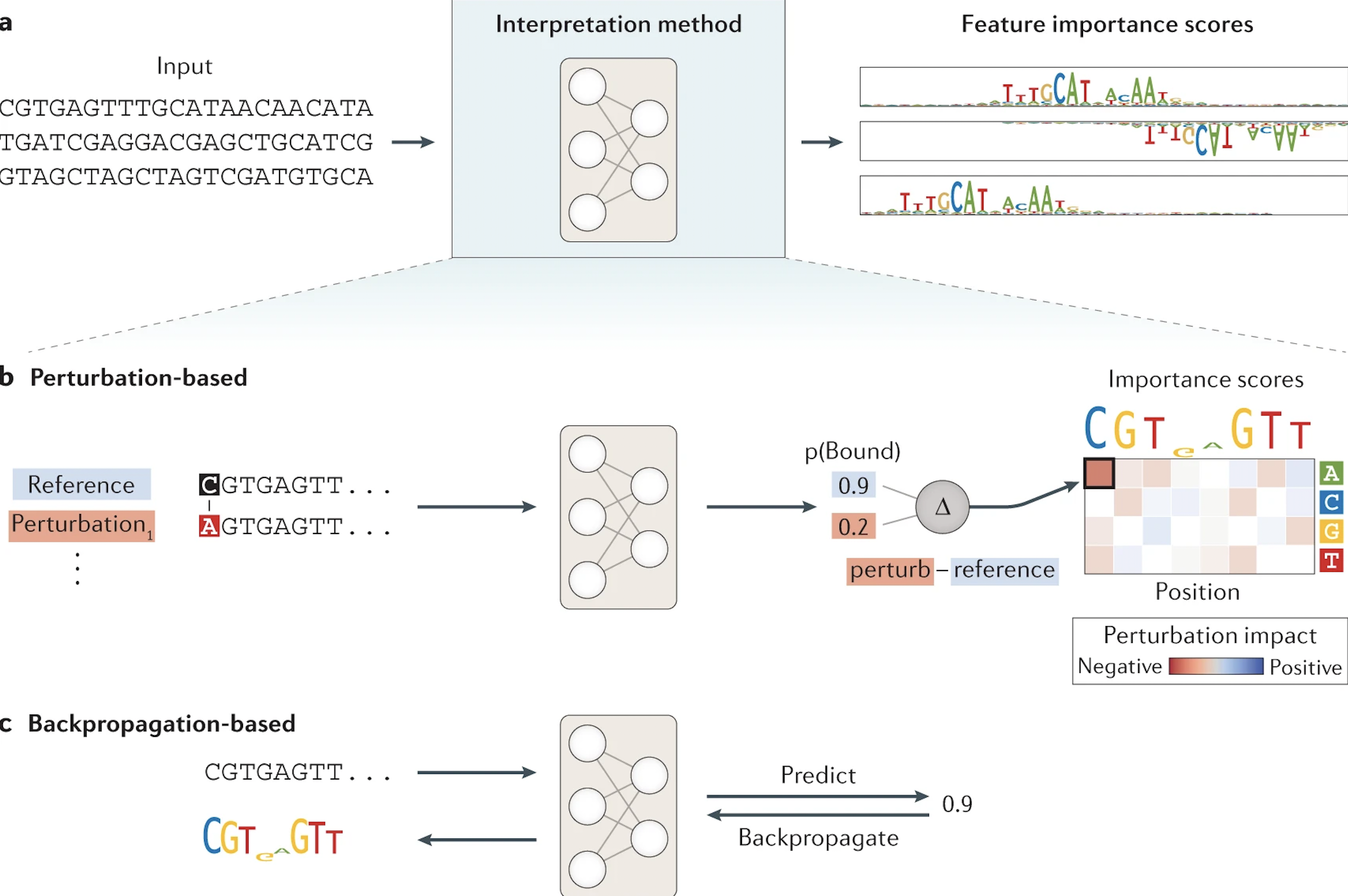}
  \caption{Different paradigms for providing explanations in AI for genomic data predictions. a) Feature importance scores proportional to letter heights state which bases of the sequence were more influential. Obtaining these scores can be done through b) Perturbation-based approaches with single base changes and c) Backpropagation-based approaches using gradients with methods such as DeepLIFT \cite{novakovsky2022obtaining}.}
  \label{fig:Genome}
\end{figure}

Attempts have been made to introduce feature importance methods like permutation approaches or gradient backpropagation to deep learning models for genomics and thus produce attribution scores which can highlight the parts of a given input that are most influential for the model prediction. In DNA sequencing this amounts to highlighting which specific nucleotides and changes affect the model prediction the most as illustrated in Figure \ref{fig:Genome}, and in most cases, they correspond to instance-level explanations for each sequence input respectively. Using saliency maps in these cases as recognised by \cite{novakovsky2022obtaining} leads to neuron saturation. If there are two identical patterns in the data such as a particular motif being repeated, erasing one would not affect the model prediction. In the case of perturbation-based gradients or input-masked gradients, the importance scores would be low for both motifs, as they are individually not necessary for the prediction. DeepLIFT and integrated gradients solve this problem by comparing the input features with their ‘reference’ values such as a shuffled version of the original sequence. As in most cases, therefore, several explainability methods should be investigated when developing XAI applications in healthcare and their limitations and results respectively compared. 

\subsubsection{EHR}
In the field of clinical risk prediction with AI, barring imaging, electronic health records (EHR) data remain the most popular source of big data in healthcare. With rising rates of digitisation across the health services sectors globally, advanced machine learning models for disease prediction can finally be developed and evaluated on more or less longitudinal, large, complex, and real-world patient data. Decades ago, promises were made by leading international organisations in the revolution of patient care provided by clinical information systems part of which EHRs would play a critical role \cite{abul2019personalized}. In the UK, more recently, EHRs played an important role in COVID-19 studies with millions of patients integrated into big data models and thus informing public guidance on health services \cite{opensafely2020opensafely}. For clinical risk prediction, EHRs remain an indispensable source of information to track patient origins, progress, and departure from clinical care alongside a larger overview of disease progression and treatment than one would usually get from discrete slides or images. 

EHR data usually consists of patient information like demographics, medical and surgical history, allergies and medications, diagnoses and procedures, vital sign measurements (heart rate, temperature, etc.), and laboratory blood test results. Some of these might be missing depending on the specific hospital wards the data originates from. For example, if working on intensive care unit (ICU) data, more regularly sampled clinical features would be present like vital signs and blood test measurements, but if operating on primary care data like longitudinal physician or clinician sources (like general practitioners in the UK), then one effectively works with static demographic and anamnesis variables with the occasional physiological measurement. The data can also be relatively structured and unstructured, sometimes requiring text-processing and natural language processing methods to extract codified diagnoses and treatment information from clinical notes \cite{pendergrass2019using}. Besides inherent complexities, one should not make the mistake of assuming EHR data is solely focussed on generating information for research. In fact, at least in the case of the US, most EHR data is actually collected for hospital billing purposes which presents later challenges in the processing pipeline like inaccuracies, missingness, and bias \cite{mahmoudi2020use}. In clinical risk prediction, this usually amounts to detailing specific attributes of the data like sampling rate in case of signal measurements, amounts of missingness and reproducible and sensible approaches to addressing the same (like with simple or advanced imputation methods), and apparent sources of bias such as patient characteristics, income distributions, and socio-racial diversity. 

\begin{figure}[h]
  \centering
  \includegraphics[width=0.5\linewidth]{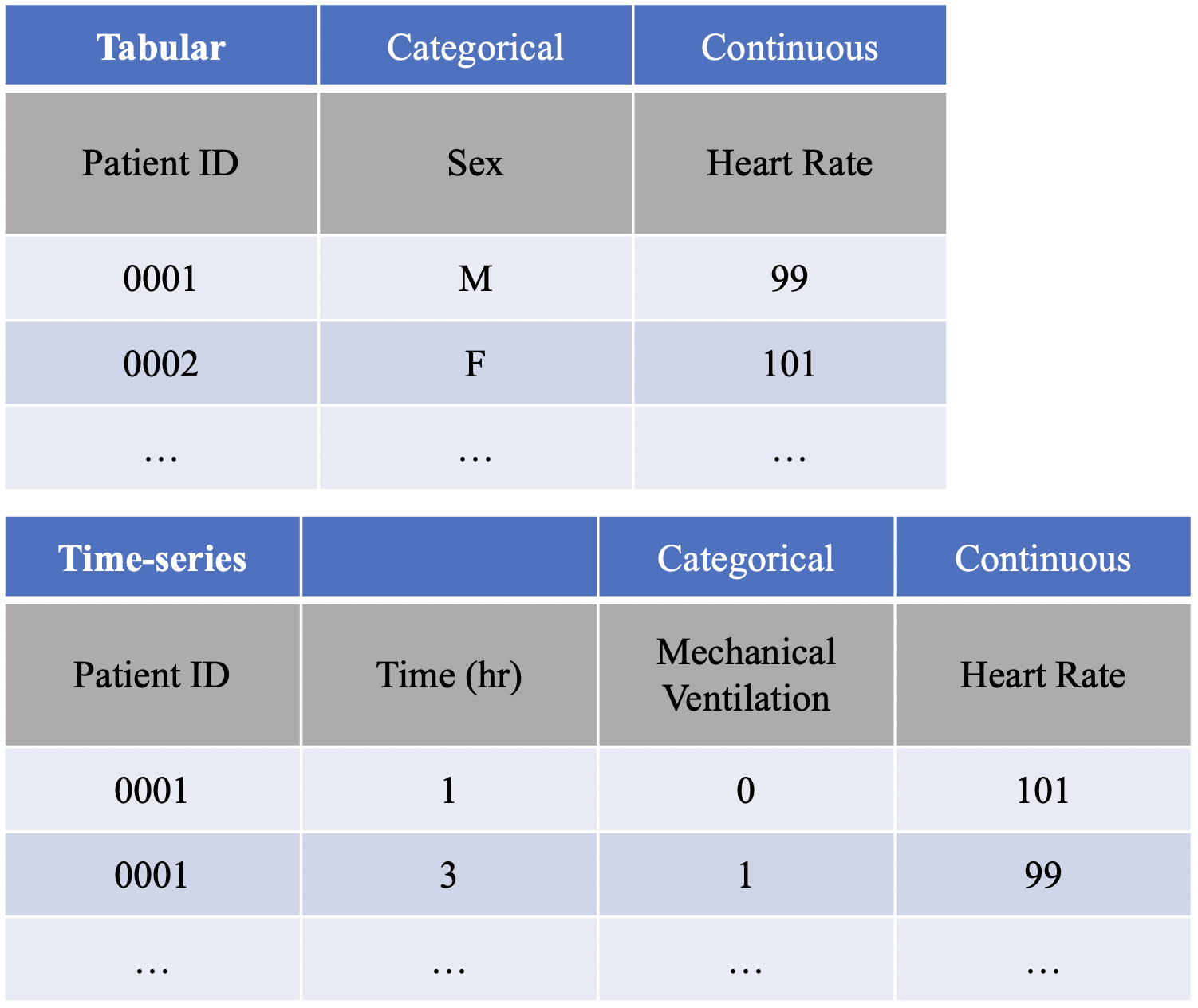}
  \caption{Two most common data formats for EHR include tabular (top) and time-series (bottom) data. Usually, one will have several measurements over time for one patient implying multi-indexing with both the patient identifier and the time-stamp for each feature for each sample in the data handling stages.}
  \label{fig:EHR}
\end{figure}

Furthermore, EHR data, as defined here, is often of a mixed nature consisting of two main modalities: static or tabular and time-series data as seen in Figure \ref{fig:EHR}. Static or tabular data usually consists of numeric or categorical variables such as patients' age, sex, ethnicity, and comorbidities, and they can usually be one-hot encoded in the case of multi-level variables. Time-series data usually refers to numeric measurements captured over time instead of just having an instantaneous single measurement for a feature. When working with general time-series data that is neither ECG nor photoplethysmography (PPG) signal, different pre-processing techniques exist. Either features are extracted like the extremes, mean, variance or standard deviation and then treated as static variables used as model input or special recurrent models are used to process the time-series for machine learning. In the case of the latter, examples include using long-short-term-memory (LSTM) units or recurrent neural networks or gated recurrent units (GRUs) to capture the long from characteristics of a time-varying feature \cite{ebrahimi2020review}. In sum, EHR records present a myriad of unique characteristics which makes the models proposed for their analysis similarly diverse. A more detailed discussion of these approaches is included in the following section detailing the studies and their findings. 

\section{Applications of XAI in Health Across Modalities}

\subsection{Imaging}

In this section, we will describe the broad strokes of XAI progress in medical imaging applications for risk prediction. Table \ref{tab:Imaging} contains a detailed listing of the identified publications. We see that the vast majority of models are based on deep learning, mostly CNN models with some autoencoder structures as well. In terms of XAI methods, there is quite a variety of applications ranging from attention, SHAP/LIME, and CAM and its extensions. Interestingly, compared to the other modalities, medical imaging applications are the most likely to be clinically validated with a team of clinicians on board for added analysis of the explanations. In terms of quantitative evaluation, a few have resorted to statistical tests to account for sampling bias but no significant evaluations were found to add legitimacy to the explanations or the applications of the methods. Reproducibility and open-access standards are lacking as a majority of the references do not include a clear link to working code-hosting services. As medical imaging is usually collected as slides by specific procedures, the dataset sizes are a lot smaller and consist of up to thousands of samples maximum and can be in the low hundreds. Some methods like Grad-CAM being able to produce relatively robust explanations across both of these extremes showcases the variety of explainability methods available right now but, sadly, insufficiently evaluated otherwise. 

\begin{table}[htp]
  \caption{XAI for clinical risk prediction for medical images for different methods, as well as evaluation criteria they might satisfy. C.V. stands for clinical validation, ie. whether the explainability was at all clinically evaluated, Q.E. for quantitative evaluation of the explainability process, and O.A. whether the application is open access}
  \label{tab:Imaging}
  \centering
\begin{tabular}{lcccccccc}
\toprule
\multirow{2}{*}{Reference} & \multirow{2}{*}{Year}  & \multicolumn{2}{c}{Problem} & \multicolumn{1}{c}{Dataset} & \multicolumn{4}{c}{XAI} \\
\cmidrule(r){3-4}
\cmidrule(r){5-6}
\cmidrule(r){6-9}
                       &       & Task  & Model   & Size            & Method          & C.V. & Q.E. & O.A.    \\
\cmidrule(r){1-9}
\cite{morvan2020learned}   & 2020               & myeloma     & CBAM     & 152      & attention  &   X &   X & X \\
\midrule
\cite{puyol2020interpretable}   & 2020               & cardiomyopathy     & VAE      & 10,000      & concepts    &   \checkmark &   X & X \\
\midrule
\cite{sabol2020explainable}   & 2020               & colorectal cancer    & CNN    & 5,000      & fuzzy    &   \checkmark & X &   X \\
\midrule
\cite{magesh2020explainable}   & 2020               & Alzheimer's     & CNN     & 642       & LIME   &   \checkmark &    X &  X\\
\midrule
\cite{binder2021morphological}   & 2021               & breast cancer     & SVM     & $>$1000      & LRP   &   \checkmark &  X &  \checkmark \\
\midrule
\cite{zhang2021whole}   & 2021               & Alzheimer's    & RL     & 1,349      & attention    &   \checkmark  & X & X   \\
\midrule
\cite{afshar2021mixcaps}   & 2021               & lung cancer    & capsule network     & 1,018      & correlation   &   \checkmark &  X &  X  \\
\midrule
\cite{wang2021icovid}   & 2021               & COVID-19 recovery    & CNN     & 2,530       & FSR   &   X & \checkmark  & \checkmark  \\
\midrule
\cite{jia2021interpretable}   & 2021               & COVID-19 mortality    & XGBoost    & 3,028       & SHAP  &   \checkmark &  X & X  \\
\midrule
\cite{liu2021predicting}   & 2021               & MVI     & CNN      & 309       & Grad-CAM    &   \checkmark &  X & X  \\
\midrule
\cite{qian2021prospective}   & 2021               & breast cancer     & CNN   & 10,815      & Grad-CAM   &   \checkmark &    X & X\\
\midrule
\cite{chang2021explaining}   & 2021               & glaucoma    & CNN    & 6,430      & adversarialism    &   \checkmark & \checkmark &   X   \\
\midrule
\cite{ballester2021predicting}   & 2021               & brain age    & CNN      & 2,639       & regression   &   \checkmark & X & \checkmark     \\
\midrule
\cite{pierson2021algorithmic}   & 2021               & osteoarthritis     & CNN    & 4,796       &  masking/CAM  &   \checkmark & \checkmark  & \checkmark   \\
\midrule
\cite{yamashita2021deep}   & 2021               & colorectal cancer     & CNN     & 343      & regression    &   \checkmark &  \checkmark &  X  \\
\midrule
\cite{le2021machine}   & 2021               & lung cancer     & XGBoost       & 211       & SHAP    &   X &   X & X   \\
\midrule
\cite{fremond2023interpretable}   & 2022               & endometrial cancer      & CNN       & 2,751       & attention    &   \checkmark &   X & \checkmark   \\
\bottomrule
\end{tabular}
\end{table}

\subsection{Text}
As we already mentioned, progress on XAI in clinical text analysis has been limited but key applications must be mentioned. Some of these are included in Table \ref{tab:Text} with a detailed description of the key characteristics of each publication article. Text modality is the most common modality present in XAI for clinical risk prediction, mostly due to the presence of medical coding applications which we take as phenotyping prediction problems. A vast array of models have been applied to these problems ranging from deep learning models, through random forests, to regression models. The vast majority of XAI techniques are either inherent due to the usage of IIM or attention which was first proposed for textual data anyway. Dataset sizes also vary in orders of magnitude, and applications include anything from ICU outcome prediction to mental health risk prediction. Most applications did not include clinical validation or quantitative evaluation for the explainability methods and few had reproducible and easily accessible code repositories for their work. We see that the pattern present in the other modalities is also present here with textual data albeit at a larger scale due to there being a lot more applications to analyse. 

\begin{table}[htp]
  \caption{XAI for clinical risk prediction for text data for different methods, as well as evaluation criteria they might satisfy. C.V. stands for clinical validation, ie. whether the explainability was at all clinically evaluated, Q.E. for quantitative evaluation of the explainability process, and O.A. whether the application is open access}
  \label{tab:Text}
  \centering
\begin{tabular}{lcccccccc}
\toprule
\multirow{2}{*}{Reference} & \multirow{2}{*}{Year}  & \multicolumn{2}{c}{Problem} & \multicolumn{1}{c}{Dataset} & \multicolumn{4}{c}{XAI} \\
\cmidrule(r){3-4}
\cmidrule(r){5-6}
\cmidrule(r){6-9}
                       &       & Task  & Model      & Size     & Method          & C.V. & Q.E. & O.A.    \\
\cmidrule(r){1-9}
\cite{girardi2018patient}   & 2019               & deterioration     & CNN        & 600,000       & attention    &   X &  X & X    \\
\midrule
\cite{kongburan2019enhancing}   & 2019               & mortality     & regression     & 23,310      & correlation   &   X &  \checkmark & X    \\
\midrule
\cite{korach2019unsupervised}   & 2019               & rapid response     & Cox      & 776,849 notes       & inherent    &   X &   \checkmark & X   \\
\midrule
\cite{mahajan2019combining}   & 2019               & readmission     & regression       & 136,963       & inherent   &   X & X&X      \\
\midrule
\cite{sun2019early}   & 2019               & AKI     & regression        & 16,558      & inherent   &   X &  X & X    \\
\midrule
\cite{zhang2019use}   & 2019               & medical screening    & regression         & 27,665       & inherent    &  X &  \checkmark & X    \\
\midrule
\cite{danielsen2019predicting}   & 2019               & mechanical restraint     & random forest     & 5,050     & inherent    &   \checkmark &   X & X\\
\midrule
\cite{miled2020predicting}   & 2020               & dementia     & random forest         & 13,747      & inherent    &   X & X&X      \\
\midrule
\cite{gong2020prognosis}   & 2020               & heart failure     & RNN       & 4,682      & attention    &   X &  X & X    \\
\midrule
\cite{bacchi2020prediction}   & 2020               & LOS     & regression        & 313       & inherent    &  X &  X & X    \\
\bottomrule
\end{tabular}
\end{table}

\begin{table}[htp]
  \caption{XAI for clinical risk prediction for text data for different methods, as well as evaluation criteria they might satisfy (continued). C.V. stands for clinical validation, ie. whether the explainability was at all clinically evaluated, Q.E. for quantitative evaluation of the explainability process, and O.A. whether the application is open access}
  \label{tab:Text}
  \centering
\begin{tabular}{lcccccccc}
\toprule
\multirow{2}{*}{Reference} & \multirow{2}{*}{Year}  & \multicolumn{2}{c}{Problem} & \multicolumn{1}{c}{Dataset} & \multicolumn{4}{c}{XAI} \\
\cmidrule(r){3-4}
\cmidrule(r){5-6}
\cmidrule(r){6-9}
                       &       & Task  & Model        & Size     & Method          & C.V. & Q.E. & O.A.    \\
\cmidrule(r){1-9}
\cite{chen2020artificial}   & 2020               & SSI     & CNN          & 21,611       & attention    &   X&  X &X    \\
\midrule
\cite{chen2020early}   & 2020               & LOS     & TF-IDF          & 12,962       & inherent   &   X &  \checkmark & X    \\
\midrule
\cite{fernandes2020predicting}   & 2020               & ICU admission     & regression*        & 120,649       & inherent   &   X &   X &X   \\
\midrule
\cite{fernandes2020risk}   & 2020               & mortality     & XGBoost       & 235,826       & inherent    &   X &  X & X    \\
\midrule
\cite{hane2020predicting}   & 2020               & dementia     & LightGBM        & 207,416       & inherent   &   X &   X& X\\
\midrule
\cite{izquierdo2020clinical}   & 2020               & ICU admission     & decision tree        & 10,504       & inherent    &   \checkmark &  X& X   \\
\midrule
\cite{korach2020mining}   & 2020               & deterioration     & Cox         &  61,740       & inherent    &  X &  X &X    \\
\midrule
\cite{li2020inferring}   & 2020               & multiple     & Bayesian latent topic         & 80,000      & inherent    &   \checkmark &  \checkmark &  \checkmark  \\
\midrule
\cite{roquette2020prediction}   & 2020               & ED admission     & CatBoost         & 499,853       & inherent   &   X &  X &  X  \\
\midrule
\cite{sterckx2020clinical}   & 2020               & preterm birth    & CatBoost       & 3,611       & inherent/SHAP    &   X &  X & \checkmark    \\
\midrule
\cite{topaz2020home}   & 2020               & ED admission    & random forest        & 89,459       & inherent    &   X &  X & X  \\
\midrule
\cite{weegar2020using}   & 2020               & cervical cancer     & random forest       & 1,321       & inherent    &   X & \checkmark & X    \\
\midrule
\cite{oliwa2021development}   & 2020               & adherence     & regression*        & 791      & inherent   &   \checkmark &   X & X \\
\midrule
\cite{dong2021explainable}   & 2021               & coding     & Bi-GRU        & 36,998       & attention    &   X &  \checkmark & X  \\
\midrule
\cite{hu2021explainable}   & 2021               & coding     & CNN      & 36,998      & attention   &   X &  X & X    \\
\midrule
\cite{barber2021natural}   & 2021               & readmission     & XGBoost        & 291      & inherent    &   X & X & X     \\
\midrule
\cite{levis2021natural}   & 2021               & suicide     & regression       & 1,232       & inherent   &   \checkmark & X & X     \\
\midrule
\cite{klang2021predicting}   & 2021               & ICU admission    & XGBoost        & 412,858      & MI    &   X &  X&  X  \\
\midrule
\cite{yang2021multimodal}   & 2021               & mortality    & LSTM       & 50,000      & attention    &  X &  X & X    \\
\midrule
\cite{kim2022can}   & 2022               & coding    & CNN       & 52,729       & attention   &   \checkmark &  \checkmark & X    \\
\midrule
\cite{lopez2023explainable}   & 2022               & coding    & transformer      & 1,300      & attention    &  X &  X & X    \\
\midrule
\cite{ahmed2022eandc}   & 2022               & mental health    & Bi-LSTM       & 15,044       & attention   &   X &  X & X    \\
\midrule
\cite{uddin2022deep}   & 2022               & depression     & LSTM      & 277,552       & LIME    &   X &  X  &  X\\
\midrule
\cite{qu2022using}   & 2022               & heart disease    & EBM          & 5,390      & inherent    &   \checkmark &   X &  X\\
\bottomrule
\end{tabular}
\end{table}

\subsection{Genomics}
Genomics is by far the least represented of the modalities in XAI for clinical risk prediction as most of the work concerns pattern classifications or dimensionality analysis and not disease prediction per se. Table \ref{tab:Genomic} shows the few papers that have done so for mortality, phenotyping, and COVID-19 severity prediction. The vast majority use some neural network architectures or IIM approaches like regressions. One of the positives is that most of these applications are indeed openly accessible with clear guidelines on how to access the code from the main paper. Some quantitative evaluation has been done including statistical tests and in cases of regression models, using external post-hoc methods to verify what has been claimed by extracting the regression coefficients. The most used post-hoc interpretability method is LIME and the dataset sizes vary considerably whether they be gene pairs or protein encodings. As far as clinical validation is concerned, most applications have not consulted clinicians or medical literature on the findings of the explainability results and that remains a potential avenue of further contribution for future applications. 

\begin{table}[htp]
  \caption{XAI for clinical risk prediction for genomic data for different methods, as well as evaluation criteria they might satisfy (continued). C.V. stands for clinical validation, ie. whether the explainability was at all clinically evaluated, Q.E. for quantitative evaluation of the explainability process, and O.A. whether the application is open access}
  \label{tab:Genomic}
  \centering
\begin{tabular}{lcccccccc}
\toprule
\multirow{2}{*}{Reference} & \multirow{2}{*}{Year}  & \multicolumn{2}{c}{Problem} & \multicolumn{1}{c}{Dataset} & \multicolumn{4}{c}{XAI} \\
\cmidrule(r){3-4}
\cmidrule(r){5-6}
\cmidrule(r){6-9}
                       &       & Task  & Model        & Size     & Method          & C.V. & Q.E. & O.A.    \\
\cmidrule(r){1-9}
\cite{kavvas2020biochemically}   & 2020               & AMR    & regression          & 1,595      & inherent   &   \checkmark &  \checkmark & \checkmark    \\
\midrule
\cite{sun2020genome}   & 2020               & mortality   & neural network          & 7,803      & LIME   &  X & \checkmark  & X    \\
\midrule
\cite{dey2021impact}   & 2021               & COVID-19 severity     & regression          & 12,965     & inherent    &   \checkmark&  \checkmark &X    \\
\midrule
\cite{van2021gennet}   & 2021               & phenotyping    & neural network          & 11,214      & inherent   &  X &  X & \checkmark    \\
\midrule
\cite{yan2021genome}   & 2021               & macular degeneration    & neural network          & 32,215      & LIME   &  X &  X & \checkmark    \\
\midrule
\cite{zhao2021deepomix}   & 2021               & mortality   & neural network          & 3,431     & inherent   &  X & X & \checkmark   \\
\bottomrule
\end{tabular}
\end{table}

\subsection{EHR}

As far as EHR applications of explainable AI are concerned, most applications rely on tabular or static data with a smaller amount of risk prediction problems dealing with time-series or sequential data. Overall, we can see that XGBoost and Shapley values are the most popular model and interpretability methods used with a diverse range of dataset sizes ranging from a few hundred to hundreds of thousands of samples. In cases of time-series measurements, attention and deep learning adaptations of Shapley values are the most common implementations of interpretability in clinical risk prediction. To avoid inflating references, we did not include research papers that use the same model and interpretability method as the ones already included to focus more on the overall trend and not include as many references as possible for their own sake. In terms of clinical tasks addressed, we see that mortality, COVID-19 outcomes (especially since 2020), and acute conditions like AKI and stroke are the most common application problems. Most of the papers did not clinically validate the interpretability results to guarantee consistency with medical literature to a significant extent, hence only including the interpretability plots or rankings as more of an afterthought rather than an end in and of itself. This stimulates the research culture in clinical risk prediction where interpretability is seen as a post-hoc step for checkmark purposes and not a significant aspect of the experimental design contributing to the research aims of the investigation at hand. Lastly, the vast majority of the papers have not made their code easily accessible from the main page either of the journal publication or the .pdf of the paper which undermines the open-access culture of the field. As compared to genomics and multi-modality sections of papers, this percentage of easily accessible code implementations is a lot smaller in the EHR case. 

As can be seen from Table \ref{tab:EHR}, most research applications in clinical risk prediction do not present any form of quantitative evaluation for their applications of interpretability methods when claiming explainability. Those that have been marked as having done so usually have only consulted their on-board clinical experts whereas only a few have more systematically used clinical feedback in evaluating the important results from their applied interpretability methods. That remains the standard in the field of clinical AI for risk prediction using EHR records especially as it concerns any attempts at quantitative evaluation which none of the included papers have attempted to do with their interpretability results. This pattern stems as a consequence partially of the lack of research impetus on such requirements in these applications as well as a lack of existing reliable methods to quantitatively evaluate interpretability methods themselves. Very recent work by \cite{turbe2023evaluation} highlights that in high-risk fields that are highly regulated like healthcare, there need to be robust quantitative evaluation frameworks for explainability of AI. Usual approaches such as using humans as evaluators, occluded datasets, and retraining models are costly, unreliable, and in some cases suffer under distribution shift. The authors propose corrupting the samples at increasing increments with the top-k and bottom-k important features as identified by the interpretability method. The change in the predictive score for each of these combinations is then plotted as two curves, one for the top-k features being corrupted, and another for the bottom-k. In a reliable interpretability method, the top-k curve measuring change in predicted score should rise dramatically as the most important features are corrupted and hence the prediction score should change significantly more than with the bottom-k curve. An AUROC and modified F1 score can be used to measure optimal behaviour between these curves and then used to compare interpretability methods. Future work will hopefully extend these results and check whether indeed, as the authors find on their few real-world and synthetic datasets, Shapley approaches remain superior methods for interpretability, at least in EHR time-series applications.

\begin{table}[htp]
  \caption{XAI for clinical risk prediction for EHR data for different methods, as well as evaluation criteria they might satisfy. C.V. stands for clinical validation, ie. whether the explainability was at all clinically evaluated, Q.E. for quantitative evaluation of the explainability process, and O.A. whether the application is open access}
  \label{tab:EHR}
  \centering
\begin{tabular}{lcccccccc}
\toprule
\multirow{2}{*}{Reference} & \multirow{2}{*}{Year}  & \multicolumn{2}{c}{Problem} & \multicolumn{1}{c}{Dataset} & \multicolumn{4}{c}{XAI} \\
\cmidrule(r){3-4}
\cmidrule(r){5-6}
\cmidrule(r){6-9}
                       &       & Task  & Model        & Size     & Method          & C.V. & Q.E. & O.A.    \\
\cmidrule(r){1-9}
\cite{elshawi2019interpretability}   & 2019               & hypertension & random forest         & 23,095    & SHAP/LIME   &  X &  \checkmark & X    \\
\midrule
\cite{pang2019understanding}   & 2019               & obesity    & XGBoost         & 860,510     & SHAP   &  X &  X & X    \\
\midrule
\cite{lu2020explainable}   & 2020               & COVID-19   & XGBoost         & 485     & SHAP   &  X&  X & X    \\
\midrule
\cite{dissanayake2020robust}   & 2020               & heart anomaly  & CNN         & 220,188     & SHAP   &  X &  X & X    \\
\midrule
\cite{song2020cross}   & 2020               & AKI & XGBoost         & 153,821   & inherent   &  X &  X & \checkmark    \\
\midrule
\cite{barda2020qualitative}   & 2020               & mortality & random forest         & -   & SHAP   &  X &  X & X    \\
\midrule
\cite{penafiel2020predicting}   & 2020               & stroke & Dempster-Shafer        & 27,876  & inherent   &  \checkmark &  \checkmark & X    \\
\midrule
\cite{hatwell2020ada}   & 2020               & multiple & AdaBoost        & -  & WHIPS   &  X &  \checkmark & \checkmark    \\
\midrule
\cite{lauritsen2020explainable}   & 2020               & acute illness & TCN     & 3,764  & DTD   &  X &  X & \checkmark    \\
\midrule
\cite{beebe2021efficient}   & 2021               & dementia    & XGBoost          & 9,103     & SHAP    &   X &  X &X    \\
\midrule
\cite{kavvas2020biochemically}   & 2021               & Alzheimer's    & LGP          & 172      & inherent   &   X &  \checkmark & \checkmark    \\
\midrule
\cite{kim2021interpretable}   & 2021               & stroke    & LightGBM          & 3,213      & SHAP   &  \checkmark &  X & X    \\
\midrule
\cite{rashed2021clinically}   & 2021               & CKD    & random forest          & 400     & SHAP   &  \checkmark &  \checkmark & X    \\
\midrule
\cite{zeng2021explainable}   & 2021               & post-op complications    & XGBoost         & 2,858     & SHAP   &  X &  X & X    \\
\midrule
\cite{zhang2021explainable}   & 2021               & AKI   & XGBoost         & 894     & SHAP   &  \checkmark &  X & X    \\
\midrule
\cite{pal2021pay}   & 2021               & ECG changes   & TabNet         & 150     & attention   &  \checkmark &  X & X    \\
\midrule
\cite{alves2021explaining}   & 2021               & COVID-19 & random forest         & 5,644    & SHAP/LIME   &  X &  \checkmark & X    \\
\midrule
\cite{moncada2021explainable}   & 2021               & mortality & XGBoost         & 36,658    & SHAP   &  X &  X & X    \\
\midrule
\cite{duckworth2021using}   & 2021               & ED admission & XGBoost         &82,402   & SHAP   &  \checkmark &  \checkmark & X    \\
\midrule
\cite{antoniadi2021prediction}   & 2021               & QOL & XGBoost         &186    & SHAP   &  \checkmark &  X & X    \\
\midrule
\cite{ward2021explainable}   & 2021               & ACS & XGBoost         &278,608    & SHAP/LIME   &  X &  \checkmark & X    \\
\midrule
\cite{zhang2021explainability}   & 2021               & PPG abnormality & CNN        & 3,764  & attention   &  X &  \checkmark & X    \\
\midrule
\cite{thimoteo2022explainable}   & 2022               & COVID-19 & XGBoost         & 1,500    & SHAP   &  X &  X & X    \\
\bottomrule
\end{tabular}
\end{table}

\subsection{Multi-modalities}

Multi-modal data is usually a combination of either text and image data or in some cases EHR and image, or EHR and text data. Few have used genomics in combination with the other modalities in explainable AI applications to clinical risk and this area presents a major potential field of growth. The models that get respectively applied are often related to the modalities, LSTMs in cases of time-series and text inclusion, CNNs if images are involved, and even XGBoost when text can be successfully summarised into a tabular format. The XAI methods consist of a plethora of mostly attention mechanisms and CAM when it comes to image data. There are examples of clinical validation where physicians were directly involved in the projects but most do not include sufficient analysis of the medical implications of the explainability results and quantitative evaluations of the XAI applications are even more limited. There are cases where open access approaches have been followed in terms of code transparency but those remain, sadly, a minority. Interestingly, attention seems to be the most popular way explainability is being incorporated into multi-modal applications for clinical risk prediction but it is still early days and this could be a limitation of existing deep learning methods applied in multi-modality applications being a more natural fit for attention (like LSTMs) for example. 

\begin{table}[htp]
  \caption{XAI for clinical risk prediction for multi-modal data for different methods, as well as evaluation criteria they might satisfy (continued). C.V. stands for clinical validation, ie. whether the explainability was at all clinically evaluated, Q.E. for quantitative evaluation of the explainability process, and O.A. whether the application is open access}
  \label{tab:Multi-modal}
  \centering
\begin{tabular}{lcccccccc}
\toprule
\multirow{2}{*}{Reference} & \multirow{2}{*}{Year}  & \multicolumn{2}{c}{Problem} & \multicolumn{1}{c}{Dataset} & \multicolumn{4}{c}{XAI} \\
\cmidrule(r){3-4}
\cmidrule(r){5-6}
\cmidrule(r){6-9}
                       &       & Task  & Model        & Size     & Method          & C.V. & Q.E. & O.A.    \\
\cmidrule(r){1-9}
\cite{hao2019page}   & 2019               & mortality    & CNN          & 2220      & inherent   &   \checkmark &  X & X    \\
\midrule
\cite{li2020inferring}   & 2020               & multiple     & Bayesian latent topic         & 80,000      & inherent    &   \checkmark &  \checkmark &  \checkmark  \\
\midrule
\cite{venugopalan2021multimodal}   & 2021               & Alzheimer's    & CNN          & 2,220     & occlusion    &   X &  X & \checkmark    \\
\midrule
\cite{ren2021interpretable}   & 2021               & pneumonia    & Bayesian network          & 35,389      & inherent   &   X &  \checkmark & X    \\
\midrule
\cite{zhang2021whole}   & 2021               & Alzheimer's    & RL         & 1,349      & attention   &   X & X & \checkmark    \\
\midrule
\cite{qian2021prospective}   & 2021               & breast cancer    & CNN         & 10,815     & CAM   &  \checkmark & X & X    \\
\midrule
\cite{barber2021natural}   & 2021               & post-op complications    & XGBoost         & 291     & inherent   &  X& X & X    \\
\midrule
\cite{yang2021multimodal}   & 2021               & mortality    & LSTM       & 50,000      & attention    &  X &  X & X    \\
\midrule
\cite{chen2022pan}   & 2022               & multiple cancers    & AMIL/SNN       & 6,592     & attention    &  \checkmark &  X & \checkmark   \\
\bottomrule
\end{tabular}
\end{table}

\section{Challenges and Future Outlook}

It is important to note that many of the included methods are, in fact, practically limited. They are sometimes vulnerable to failures, redundant explanations, and wrong indications which makes interpretability on its own an insufficient attribute for achieving explainable and reliable AI clinical risk models \cite{ghassemi2021false}. Some work like \cite{hatwell2020ada, duckworth2021using} and others have in recent years addressed the need for external validation if not as a companion to increased explainability then in lieu of it. These strategies for XAI in healthcare applications should be taken with a grain of salt as models externally validated on similar patient cohorts might not add additional trust or evaluate fairness more robustly than intuitive explainability methods. Ideally, a combination of both external validation with diverse stratified sub-population cohorts based on socio-economic and comorbidity groups and robust interpretability methods whether they be inherent or post-hoc is recommended to achieve explainability. An important extension should be to develop more robust explainability frameworks that can work across modalities and methodological contexts, as well as provide evidence for external multi-centre validation of proposed prediction models. 

From the overview of explainable AI applications for clinical risk prediction across multiple modalities, it is clear that only rarely have the benefits of the models or explanations been evaluated for the clinician or patient reception. Some studies have included tests conducted with clinicians and patients for the generated explanations but sadly claims of explainable AI are greatly exaggerated in most cases. Furthermore, the added performance benefits of using opaque deep learning models at the cost of explainability are not convincing and the community seems to acknowledge this in modalities such as text and EHR where the majority of models are IIM or decision-based classifiers like XGBoost. The approach taken, thus, is to not have to compromise significantly between the trade-off but rather make inherently interpretable models that can also be high-performing. It is difficult to beat XGBoost and its cousins in tabular machine learning, even with deep learning models like TabNet and NODE. This approach should be extended to other modalities as well. 

Some possible ideas for adoption of more rigorous testing of applied explainability in clinical risk prediction modelling is to test the methods on synthetic datasets with known underlying generative factors. There are already some examples of such datasets extracted from existing and in-use clinical datasets like the Clinical Practice Research Datalink (CPRD) containing EHRs of millions of patients in the United Kingdom or UK Biobank which similarly contains a large amount of relatively complex and diverse patient data \cite{herrett2015data}. Testing that the implemented XAI methods correctly identify most of the known factors is an important reliability test before being proposed to be used in clinical decision-making systems. It would add further trust when negotiating with clinicians regarding uptake. In our review of EHR time-series applications, we highlighted recent developments in implementing quantitative evaluations using a combination of corruption and ranking solutions based on adapted AUROC and F1 scores. More work still needs to be done on suggesting further frameworks for robust evaluation and comparison of interpretability or explainability more broadly in EHR and other clinical risk prediction domains. 

While explainable AI holds great potential for revolutionizing clinical risk prediction, it is essential to recognize that it is not a panacea. Human expertise and domain knowledge remain indispensable in healthcare decision-making. Explainable AI should be seen as a tool to assist and augment healthcare professionals' judgment, providing them with transparent insights into the underlying factors contributing to predictions. If researchers seek to use explainability to guarantee the trust and reliability of their models for clinicians, patients, and other stakeholders, their implementations not being transparent or easily accessible sends the wrong message and stifles research growth. Since the papers that do have open access links on their publications are from high-ranking journals that demand open access and code sharing when submitting, a possible way to mitigate this is to have more journals demand accessible code sharing resources from authors, especially in cases of explainability research which, as we have seen, cannot be separated from concepts around trustworthiness, the key to which is transparency and reproducibility in the case of software. 

We should not, however, be overly critical or pessimistic about the outcomes of explainability research in clinical risk prediction. The medical field is more than familiar with using black-box technology as many drugs' mechanisms have still not been elucidated for their health benefits. Paracetamol is a commonly cited example of a popular over-the-counter drug whose mechanisms of action have not been revealed for a much longer time than we have had deep learning \cite{kirkpatrick2005new, ghassemi2021false}. Several methods for achieving interpretability in clinical risk prediction were explored, including rule-based models, feature importance techniques, and post-hoc methods but a key point has been made of interpretability on its own being an insufficient attribute of a truly explainable clinical risk prediction AI model. Each approach, thus, presented its own advantages and limitations, emphasizing the need for a careful selection and integration of multiple techniques to ensure a comprehensive understanding of AI-driven predictions. The progress forward has to be focused on an end-to-end approach to explainability in clinical risk prediction, no longer being enough to simply apply an explainability method to a model, but rather to clinically and quantitatively measure its success while also including different stakeholders from clinicians, and patients, to developers into the process. Clear regulations, guidelines, and research culture practices will help make this transition smoother and for the benefit of a larger group of stakeholders.

\bibliographystyle{unsrt}  
\bibliography{references}  

\end{document}